\documentclass[10pt,twocolumn,letterpaper]{article}

\usepackage[article]{cvpr}

\usepackage{graphicx}
\usepackage{amsmath}
\usepackage{amssymb}
\usepackage{booktabs}
\usepackage{multirow}
\usepackage{array}
\usepackage{colortbl}
\usepackage{xcolor}
\usepackage{listings}
\usepackage{tikz}
\usetikzlibrary{shapes,arrows,positioning,fit,backgrounds}
\usepackage{ctable}
\usepackage{enumitem}

\definecolor{cvprblue}{rgb}{0.21,0.49,0.74}
\usepackage[pagebackref,breaklinks,colorlinks,citecolor=cvprblue,linkcolor=cvprblue,urlcolor=cvprblue]{hyperref}

\lstdefinestyle{codestyle}{
    basicstyle=\ttfamily\scriptsize,
    breaklines=true,
    frame=single,
    numbers=left,
    numberstyle=\tiny,
    tabsize=2
}

\def\paperID{}
\def\confName{CVPR}
\def\confYear{2026}

\begin{document}

\pagestyle{plain}
	
\title{Delta-Based Neural Architecture Search:\\LLM Fine-Tuning via Code Diffs}

\author{Santosh Premi Adhikari, \space\space\space Radu Timofte,\space\space\space Dmitry Ignatov\\
	\small{Computer Vision Lab, CAIDAS \& IFI, University of W\"urzburg, Germany}}
\maketitle

\begin{abstract}
Large language models (LLMs) show strong potential for neural architecture generation, yet existing approaches produce complete model implementations from scratch---a paradigm that is computationally expensive and yields verbose, redundant code. We propose \textbf{Delta-Code Generation}, a paradigm where fine-tuned LLMs generate compact unified diffs (deltas) to refine existing baseline architectures rather than synthesizing entire models. Our pipeline iteratively fine-tunes the LLM via LoRA on curated architectures from the LEMUR dataset, with MinHash-Jaccard novelty filtering to maintain structural diversity. We evaluate three 7B-class code/instruct LLMs---DeepSeek-Coder-7B-Instruct-v1.5, Qwen2.5-Coder-7B-Instruct, and Mistral-7B-Instruct-v0.3---across six datasets (CIFAR-10, CIFAR-100, MNIST, SVHN, ImageNette, and CelebA) using the same 22-cycle protocol (1,100 candidates per LLM). Under a shared cluster protocol with a balanced six-dataset baseline pool, all three LLMs yield closely matched headline metrics: DeepSeek-Coder reaches 75.3\% valid generation rate, 65.8\% mean first-epoch accuracy, and 99.5\% best first-epoch accuracy; Qwen2.5-Coder reaches 72.1\% / 64.6\% / 99.5\%; and Mistral-7B 66.6\% / 66.1\% / 99.5\%---all accuracies being first-epoch validation accuracy, the proxy used in prior LLM-based NAS work~\cite{ABrain.Architect,gu2026iterative}. All three substantially surpass the full-generation baseline (50.6\% valid rate, 42.3\% mean first-epoch accuracy, 64.0\% best first-epoch accuracy on CIFAR-10 only). On CIFAR-10 specifically, Mistral, DeepSeek, and Qwen reach 85.5\%, 85.2\%, and 80.6\% best first-epoch accuracy respectively---all well above 63.98\% full generation and 71.5\% for the concurrent iterative approach of Gu~\etal. Output lengths are uniformly compact---30.4 lines (DeepSeek), 31.4 lines (Qwen), and 49.5 lines (Mistral)---a 75--85\% reduction over full generation across all three LLMs. A 50-epoch full-training study on the top-20 architectures per LLM confirms that the 1-epoch proxy reliably preserves performance rankings (Mistral: Spearman $\rho = 0.926$), placing it at the upper end of NAS proxy correlations reported in the literature. Our results position delta-based generation as a token-efficient, multi-domain, and LLM-agnostic alternative to full-model synthesis for LLM-driven neural architecture search. All admitted architectures, per-model statistics, and fine-tuning configurations are released under the \texttt{del-} prefix in the LEMUR dataset for reviewer verification and full reproducibility.
\end{abstract}

\section{Introduction}
\label{sec:intro}

Neural Architecture Search (NAS) has emerged as a powerful paradigm for automating the design of deep neural networks~\cite{zoph2017neural,elsken2019neural,white2023nas}. Early NAS methods based on reinforcement learning required up to 22,400 GPU-days~\cite{zoph2017neural}, while even efficient differentiable approaches like DARTS~\cite{liu2019darts} depend on heavy supernet training. Standardized benchmarks such as NAS-Bench-201~\cite{dong2020nasbench201} have enabled reproducible comparisons, yet the search space remains constrained to predefined cell structures. The recent advent of code-capable LLMs~\cite{devlin2019bert,touvron2023llama,roziere2023code,guo2024deepseek,hui2024qwen25coder} has opened an alternative avenue: using language models as architecture generators that produce executable PyTorch code directly~\cite{nasir2023llmatic,cai2025seki,ABrain.Architect}.

Within this LLM-based NAS paradigm, Khalid~\etal~\cite{ABrain.Architect} demonstrated that iteratively fine-tuning an LLM on its own successful generations can progressively improve both generation reliability and model quality. Their framework generates \emph{complete} PyTorch implementations, achieving a 50.6\% valid generation rate (Wilson 95\% CI [45.0\%, 56.1\%]) and a best first-epoch accuracy of 63.98\% with mean 50.99\% at cycle~18 on CIFAR-10, after 22 fine-tuning cycles. The updated version of their work extends the evaluation to three datasets (CIFAR-10, CIFAR-100, SVHN) using dataset-specific accuracy thresholds (40\%, 20\%, 70\% respectively), and we treat it as the current state of the art for LLM-based fixed-semantics generation on the LEMUR benchmark. Concurrently, Gu~\etal~\cite{gu2026iterative} (a concurrent arXiv preprint) proposed a closed-loop iterative pipeline using frozen LLMs with historical feedback memory, achieving up to 71.5\% one-epoch accuracy on CIFAR-10 through 2000 iterations without any LLM fine-tuning.

However, both approaches share a critical limitation: the LLM must generate complete model implementations at each step. This full-generation paradigm produces verbose outputs (typically 200+ lines), requires regenerating working code structures, and cannot leverage the structural knowledge embedded in existing architectures.

We propose \textbf{Delta-Code Generation}, an approach where the LLM generates compact unified diffs (deltas) to modify existing baseline architectures rather than synthesizing complete implementations. Drawing inspiration from the software engineering practice of patch-based code modification~\cite{xia2023automated,jimenez2024swe}, our paradigm constrains the generation space to targeted architectural refinements. This constrained generation space offers several distinct advantages. Primarily, it drastically reduces token consumption; delta outputs average 30 lines for DeepSeek, 31 lines for Qwen, and 49.5 lines for Mistral---versus 200+ lines for full-model synthesis (75--85\% reduction across all three LLMs). Furthermore, the approach inherently preserves the structural integrity of working baselines while targeting specific architectural improvements. It also scales naturally across diverse domains, allowing us to evaluate generalization on six distinct datasets simultaneously rather than being limited to CIFAR-10. Finally, this framework provides a rigorous testbed for comparing modern code-oriented LLMs, such as DeepSeek-Coder and Qwen2.5-Coder, in the context of structured code generation.

In summary, our main contributions are as follows. We introduce Delta-Code Generation, a novel NAS paradigm that synthesizes unified diffs instead of complete models, cutting output length by 75--85\% across three distinct 7B LLMs while preserving competitive first-epoch proxy accuracy (evaluated under the same one-epoch protocol as prior LLM-based NAS work~\cite{ABrain.Architect,gu2026iterative}). We also provide a comprehensive head-to-head evaluation of three 7B-class LLMs---DeepSeek-Coder-7B, Qwen2.5-Coder-7B, and Mistral-7B-Instruct---across two distinct code-training lineages (code-specialized and general-instruct), highlighting the impact of recent LLM advancements on structured generation tasks and demonstrating that delta generation transfers across LLM families. Lastly, we extend the evaluation protocol beyond single-dataset benchmarks to encompass six diverse image classification datasets, demonstrating the robust multi-domain capabilities of our approach.

\section{Related Work}
\label{sec:Related}

\subsection{Neural Architecture Search}

NAS methods automate network design through various optimization strategies. Reinforcement learning-based NAS~\cite{zoph2017neural} and evolutionary algorithms~\cite{real2019regularized} achieve strong results but require thousands of GPU-hours. Parameter-sharing methods such as ENAS~\cite{pham2018enas} and differentiable approaches like DARTS~\cite{liu2019darts} reduced search cost by orders of magnitude, while hand-designed scaling rules~\cite{tan2019efficientnet} and residual connections~\cite{he2016resnet} have proven highly effective within fixed topologies. Standardized benchmarks such as NAS-Bench-201~\cite{dong2020nasbench201} enabled reproducible comparisons across methods. More recently, zero-shot NAS methods~\cite{mellor2021naswot,li2024zeroshot} bypass training entirely using proxy metrics, yet remain constrained to predefined search spaces.

\subsection{LLM-Based Architecture Generation}

LLMs have been increasingly applied to architecture design. Zheng~\etal~\cite{zheng2023can} explored GPT-4 for NAS, while AutoML-GPT~\cite{zhang2023automl} proposed end-to-end AutoML with LLMs. LLMatic~\cite{nasir2023llmatic} combines LLM-driven mutation with quality-diversity search. EvoPrompting~\cite{chen2023evoprompting} applies evolutionary prompting with LLMs as mutation operators for code-level NAS. SEKI~\cite{cai2025seki} uses self-evolution and knowledge inspiration for LLM-guided NAS, achieving competitive performance in 0.05 GPU-days. RZ-NAS~\cite{ji2025rznas} integrates reflective zero-cost strategies. LEMONADE~\cite{rahman2025automated} uses ChatGPT for multi-objective architecture discovery. Within the NNGPT framework~\cite{ABrain.NNGPT}, Khalid~\etal~\cite{ABrain.Architect} established a 22-cycle iterative fine-tuning paradigm, and Gu~\etal~\cite{gu2026iterative} proposed iterative NAS with historical feedback memory using frozen LLMs. Extensions to the NNGPT framework include few-shot prompting strategies for improved generation reliability~\cite{ABrain.Prompt}, fractal-inspired computational architectures~\cite{ABrain.NNGPT-Fractal}, discovery of non-standard channel priors~\cite{ABrain.CV_Channel}, and retrieval-augmented approaches to extract algorithmic logic from neural networks~\cite{ABrain.NN-RAG}. Separately, HPGPT~\cite{ABrain.HPGPT} demonstrated that LLMs can predict training hyperparameters, treating HP optimization as a distinct task from architecture generation. In parallel, the software engineering community has developed patch-based code generation benchmarks such as SWE-bench~\cite{jimenez2024swe}, demonstrating that LLMs can generate targeted diffs to resolve real-world issues~\cite{chen2021evaluating,liu2024survey}. Our work bridges these two directions by applying diff-based generation to neural architecture refinement, fine-tuning the LLM on successful delta generations rather than complete models.

\subsection{Positioning Relative to Prior Work}

Table~\ref{table:positioning} positions our method against the most relevant LLM-based NAS approaches. We are the only method that combines fine-tuned delta generation with multi-dataset evaluation and structural novelty filtering, offering a unique trade-off between generation efficiency and architectural diversity.

\begin{table}[t]
    \caption{Positioning relative to LLM-based NAS methods. $\Delta$: generates diffs rather than complete models; FT: fine-tunes the LLM; Multi: evaluated on $>$3 datasets; Novel: structural novelty filtering.}
    \label{table:positioning}
    \centering
    \scriptsize
    \setlength{\tabcolsep}{3pt}
    \begin{tabular}{lcccccc}
        \toprule
        \textbf{Method} & \textbf{NAS} & \textbf{Open} & \textbf{$\Delta$} & \textbf{FT} & \textbf{Multi} & \textbf{Novel} \\
        \midrule
        EvoPrompting~\cite{chen2023evoprompting} & \checkmark & \checkmark & $\times$ & $\times$ & $\times$ & $\times$ \\
        SEKI~\cite{cai2025seki} & \checkmark & \checkmark & $\times$ & $\times$ & $\times$ & $\times$ \\
        LLMatic~\cite{nasir2023llmatic} & \checkmark & \checkmark & $\times$ & $\times$ & $\times$ & \checkmark \\
        LEMONADE~\cite{rahman2025automated} & \checkmark & $\times$ & $\times$ & $\times$ & \checkmark & $\times$ \\
        Khalid~\etal~\cite{ABrain.Architect} & \checkmark & \checkmark & $\times$ & \checkmark & $\times$ & \checkmark \\
        Gu~\etal~\cite{gu2026iterative} & \checkmark & \checkmark & $\times$ & $\times$ & \checkmark & $\times$ \\
        \textbf{Ours} & \checkmark & \checkmark & \checkmark & \checkmark & \checkmark & \checkmark \\
        \bottomrule
    \end{tabular}
\end{table}

\section{Methodology}
\label{sec:Methodology}

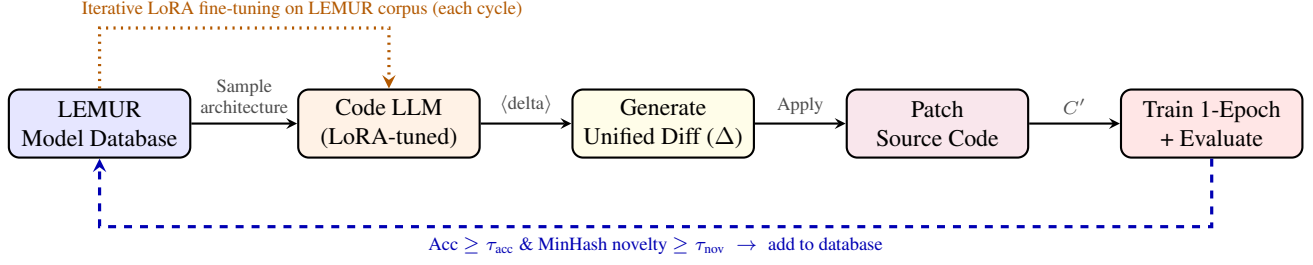
\begin{figure*}[t]
    \centering
    \begin{tikzpicture}[
        node distance=0.7cm and 1.0cm,
        box/.style={rectangle, draw, rounded corners, minimum height=0.9cm, minimum width=2.4cm, align=center, font=\small, thick},
        arrow/.style={->, thick, >=stealth},
        label/.style={font=\scriptsize, align=center, text=black!70}
    ]
        \node[box, fill=blue!10] (db) {LEMUR\\Model Database};
        \node[box, fill=orange!10, right=1.4cm of db] (llm) {Code LLM\\(LoRA-tuned)};
        \node[box, fill=yellow!10, right=1.2cm of llm] (delta) {Generate\\Unified Diff ($\Delta$)};
        \node[box, fill=purple!10, right=1.2cm of delta] (patch) {Patch\\Source Code};
        \node[box, fill=red!10, right=1.2cm of patch] (eval) {Train 1-Epoch\\+ Evaluate};

        \draw[arrow] (db) -- node[above, label] {Sample\\architecture} (llm);
        \draw[arrow] (llm) -- node[above, label] {$\langle$delta$\rangle$} (delta);
        \draw[arrow] (delta) -- node[above, label] {Apply} (patch);
        \draw[arrow] (patch) -- node[above, label] {$C'$} (eval);

        \draw[arrow, dashed, color=blue!70!black, line width=1.2pt] (eval.south) -- ++(0,-0.9) -| node[below, pos=0.25, label, color=blue!70!black] {Acc $\geq \tau_\text{acc}$ \& MinHash novelty $\geq \tau_\text{nov}$ $\;\rightarrow\;$ add to database} (db.south);

        \draw[arrow, dotted, color=orange!70!black, line width=1.2pt] (db.north) -- ++(0,0.8) -| node[above, pos=0.35, label, color=orange!70!black] {Iterative LoRA fine-tuning on LEMUR corpus (each cycle)} (llm.north);

    \end{tikzpicture}
    \caption{The Delta-Code Generation iterative pipeline. At each cycle, the LLM samples an architecture from the database, generates a targeted unified diff ($\Delta$), and the patched code is trained for one epoch. Successful, structurally novel architectures are added back to the database (dashed). The LLM is fine-tuned via LoRA on the LEMUR corpus at each cycle (dotted), progressively learning the diff format.}
    \label{fig:overview}
\end{figure*}

We propose Delta-Code Generation, a framework where LLMs generate unified diffs to modify existing neural network architectures rather than synthesizing complete implementations. Figure~\ref{fig:overview} illustrates the pipeline.

\subsection{Problem Formulation}

Given a baseline neural network architecture $\mathcal{B}$ with source code $C_\mathcal{B}$, our objective is to learn a delta generator $G_\theta$ parameterized by an LLM with parameters $\theta$ such that:
\begin{equation}
    \Delta = G_\theta(C_\mathcal{B}, \mathcal{D}), \quad C_\mathcal{B}' = \text{\textsc{Apply}}(C_\mathcal{B}, \Delta)
\end{equation}
where $\mathcal{D}$ is the target dataset descriptor and $\text{\textsc{Apply}}$ is the standard unified diff application operator. The generated architecture $C_\mathcal{B}'$ is trained for a single epoch, and the process succeeds when:
\begin{equation}
    \mathcal{A}(C_\mathcal{B}', \mathcal{D}) \geq \tau_{\text{acc}} \quad \text{and} \quad \mathcal{N}(C_\mathcal{B}', \mathcal{S}) \geq \tau_{\text{nov}}
\end{equation}
where $\mathcal{A}(\cdot)$ denotes first-epoch validation accuracy, $\mathcal{N}(\cdot)$ measures structural novelty against the existing corpus $\mathcal{S}$, $\tau_{\text{acc}} = 0.40$ is the accuracy threshold, and $\tau_{\text{nov}} = 0.90$ is the MinHash-Jaccard novelty threshold.

\subsection{Delta Output Format}

Unlike full-generation approaches that produce complete Python files, our method constrains the LLM to generate structured XML output. The primary component is the \texttt{<delta>} tag, which contains a unified diff in standard patch format with context lines, additions (+), and deletions (-).

The prompt template also elicits \texttt{<hp>} (hyperparameters) and \texttt{<tr>} (transform code) tags, following the NNGPT framework convention~\cite{ABrain.NNGPT}. Data transforms are handled through a registry-based system~\cite{ABrain.Transform} rather than LLM-generated code. Consistent with the evaluation protocol of Khalid~\etal~\cite{ABrain.Architect} and Gu~\etal~\cite{gu2026iterative}, the \emph{evaluation} of generated architectures uses the baseline model's original hyperparameters and transforms from the LEMUR database, ensuring that only architectural changes are measured. The LLM-generated \texttt{<hp>} and \texttt{<tr>} thus serve as structured context within the output format rather than direct evaluation inputs.

This structured format reduces ambiguity and enables reliable parsing. The prompt instructs the LLM to generate compact deltas (at most 30 lines), relying on prompt engineering to guide generation quality. An example is shown in Figure~\ref{fig:delta_example}.

\begin{figure}[t]
\begin{lstlisting}[style=codestyle, language=Python]
<hp>
{"batch": 64, "lr": 0.01, "momentum": 0.9}
</hp>

<tr>
import torchvision.transforms as transforms
def transform(norm):
    return transforms.Compose([
        transforms.Resize(32),
        transforms.ToTensor(),
        transforms.Normalize(*norm)])
</tr>

<delta>
--- baseline.py
+++ improved.py
@@ -21,6 +21,7 @@ class Net(nn.Module):
     def __init__(self, ...):
         super().__init__()
         self.conv1 = nn.Conv2d(3, 64, 3)
+        self.bn1 = nn.BatchNorm2d(64)
         self.fc = nn.Linear(64, 10)
</delta>
\end{lstlisting}
\caption{Example delta output. The LLM generates hyperparameters, transform code, and a unified diff that adds batch normalization---a targeted modification rather than a complete model rewrite.}
\label{fig:delta_example}
\end{figure}

\subsection{Pipeline}

Figure~\ref{fig:algorithm} formalizes the delta-code generation pipeline. Each fine-tuning cycle consists of: (1) generating candidate deltas from baseline architectures, (2) applying and validating the deltas, (3) filtering by accuracy and novelty, and (4) fine-tuning the LLM on successful generations.

\begin{figure}[t]
\small
\hrule
\vspace{2pt}
\textbf{Algorithm 1:} Delta-Code Generation Pipeline \\
\hrule
\vspace{4pt}
\textbf{Input:} LLM $G_\theta$, baselines $\mathcal{L}$, cycles $E$, candidates/cycle $N$, \\
\hspace*{1.1cm} accuracy threshold $\tau_\text{acc}$, novelty threshold $\tau_\text{nov}$ \\
\textbf{Output:} Fine-tuned LLM $G_{\theta^*}$, generated corpus $\mathcal{S}$
\vspace{2pt}
\hrule
\vspace{2pt}
\begin{tabular}{@{\hspace{0pt}}r@{\hspace{4pt}}l}
1: & $\mathcal{S} \gets \emptyset$ \\
2: & \textbf{for} $e = 1$ \textbf{to} $E$ \textbf{do} \\
3: & \quad $\mathcal{C}_e \gets \emptyset$ \\
4: & \quad \textbf{for} $i = 1$ \textbf{to} $N$ \textbf{do} \\
5: & \quad\quad Sample baseline $(\mathcal{B}_i, C_{\mathcal{B}_i}) \sim \mathcal{L}$ \\
6: & \quad\quad $\Delta_i \gets G_\theta(C_{\mathcal{B}_i})$ \\
7: & \quad\quad \textbf{if} \textsc{Apply}$(C_{\mathcal{B}_i}, \Delta_i)$ succeeds \textbf{then} \\
8: & \quad\quad\quad $C_i' \gets$ \textsc{Apply}$(C_{\mathcal{B}_i}, \Delta_i)$ \\
9: & \quad\quad\quad \textbf{if} \textsc{Validate}$(C_i')$ passes \textbf{then} \\
10: & \quad\quad\quad\quad $a_i \gets$ \textsc{Train-1-Epoch}$(C_i', \mathcal{D}_i)$ \\
11: & \quad\quad\quad\quad \textbf{if} $a_i \geq \tau_\text{acc}$ \textbf{then} \\
12: & \quad\quad\quad\quad\quad $n_i \gets$ \textsc{NoveltyScore}$(C_i', \mathcal{S})$ \\
13: & \quad\quad\quad\quad\quad \textbf{if} $n_i \geq \tau_\text{nov}$ \textbf{then} \\
14: & \quad\quad\quad\quad\quad\quad $\mathcal{C}_e \gets \mathcal{C}_e \cup \{(C_{\mathcal{B}_i}, \Delta_i, a_i)\}$ \\
15: & \quad $\mathcal{S} \gets \mathcal{S} \cup \mathcal{C}_e$ \\
16: & \quad $\theta \gets$ \textsc{LoRA-Fine-Tune}$(G_\theta, \mathcal{L})$ \\
17: & \textbf{return} $G_{\theta}$, $\mathcal{S}$
\end{tabular}
\vspace{2pt}
\hrule
\caption{Delta-Code Generation pipeline. At each cycle, the LLM generates deltas for $N$ baselines. $\mathcal{D}_i$ denotes the baseline's dataset configuration (hyperparameters and transforms from LEMUR). \textsc{NoveltyScore}$(C_i',\mathcal{S}) = 1 - \max_{C\in\mathcal{S}}\,\text{MinHash-Jaccard}(C_i',C)$ returns a structural dissimilarity in $[0,1]$, so the admission condition $n_i\!\geq\!\tau_\text{nov}$ keeps candidates that are \emph{distant} (at most $1-\tau_\text{nov}$ similar) from the current corpus. Only candidates passing delta application, validation, accuracy ($\geq\tau_\text{acc}$), and novelty ($\geq\tau_\text{nov}$) filters are added to the training corpus for subsequent LoRA fine-tuning.}
\label{fig:algorithm}
\end{figure}

\textbf{Baseline Selection.} We sample baselines from the LEMUR Neural Network Dataset~\cite{ABrain.NN-Dataset,ABrain.LEMUR2}, which provides diverse architectures with performance annotations across multiple datasets.

\textbf{Delta Generation and Application.} The LLM receives the baseline code and generates structured output. The unified diff is applied using standard patch utilities. If the patch fails (context mismatch, invalid format), the generation is marked as unsuccessful.

\textbf{Validation.} Successfully patched architectures undergo syntax validation, instantiation checking, forward pass verification, and single-epoch training.

\textbf{Filtering.} Before admitting any successfully patched architecture into the training corpus, we apply two strict filters. First, an accuracy filter discards any model that fails to achieve at least 40\% top-1 validation accuracy after a single epoch of training~\cite{ABrain.Architect}. Second, a novelty filter ensures structural diversity by computing the MinHash-Jaccard similarity~\cite{ABrain.Architect} of the generated code against all existing corpus members. Using 256 permutations and 10-character shingles, we reject any architecture that does not exhibit at least 90\% structural dissimilarity ($\geq 0.90$), preventing the dataset from collapsing into near-duplicate designs.

\textbf{Fine-Tuning.} At each cycle, the LLM is fine-tuned via LoRA~\cite{hu2022lora} on the \emph{static} LEMUR training corpus $\mathcal{L}$ (Algorithm~1, line~16). The newly generated and novelty-admitted architectures are added to the baseline \emph{sampling pool} for future delta generation (line~15), but the LoRA training data itself remains the fixed, curated LEMUR corpus throughout all 22 cycles. This design ensures a stable and diverse training signal: the LLM learns the diff format and structural patterns from high-quality LEMUR examples, while the expanding generated pool $\mathcal{S}$ increases the diversity of baselines available for modification in subsequent cycles.

\subsection{LLM Fine-Tuning Configuration}

To adapt the language models efficiently, we apply Low-Rank Adaptation (LoRA)~\cite{hu2022lora} across all attention and MLP projections (q, k, v, o, up, down, gate) as well as the \texttt{lm\_head}. We set the LoRA rank $r=32$ and scaling factor $\alpha=32$ with a dropout rate of 0.05. Training proceeds in bfloat16 precision using a learning rate of $1 \times 10^{-5}$ governed by a cosine decay schedule, 20 warmup steps, and a weight decay of 0.01. To manage memory constraints, we use a per-device batch size of 1 coupled with 8 gradient accumulation steps. During each iterative cycle, the model undergoes 3 LoRA fine-tuning epochs. For the inference phase, we sample architectures using a temperature of 0.35, top-$k$ of 50, and top-$p$ of 0.9, capping the generation at 1,024 new tokens.

\section{Experiments}
\label{sec:experiments}

\subsection{Datasets}

Unlike prior work that evaluates on CIFAR-10 alone~\cite{ABrain.Architect} or three datasets~\cite{gu2026iterative}, we assess generalization across six diverse image classification datasets spanning different resolutions, class counts, and visual domains (Table~\ref{tab:datasets}). Data augmentation follows established techniques~\cite{Aboudeshish2025augmentation}. The baselines sampled from LEMUR span all six datasets, exposing the LLM to diverse classification tasks during training.

\begin{table}[t]
    \centering
    \small
    \caption{Summary of evaluation datasets. Our pipeline handles datasets ranging from $28{\times}28$ grayscale digits to $178{\times}218$ RGB facial images---a significantly broader scope than prior LLM-based NAS work.}
    \label{tab:datasets}
    \setlength{\tabcolsep}{2.5pt}
    \begin{tabular}{lcccl}
        \toprule
        \textbf{Dataset} & \textbf{Cls.} & \textbf{Resolution} & \textbf{Train} & \textbf{Domain} \\
        \midrule
        CIFAR-10~\cite{krizhevsky2009learning} & 10 & $32{\times}32$ & 50K & Objects \\
        CIFAR-100~\cite{krizhevsky2009learning} & 100 & $32{\times}32$ & 50K & Fine-grained \\
        MNIST~\cite{lecun1998mnist} & 10 & $28{\times}28$ & 60K & Digits \\
        SVHN~\cite{netzer2011svhn} & 10 & $32{\times}32$ & 73K & Street digits \\
        ImageNette~\cite{howard2019imagenette} & 10 & $160{\times}160$ & 9.5K & ImageNet subset \\
        CelebA~\cite{liu2015celeba} & 2 & $178{\times}218$ & 163K & Facial attr. \\
        \bottomrule
    \end{tabular}
\end{table}

\subsection{Base Models}

We compare three 7B-class LLMs spanning two training lineages (code-specialized and general-purpose instruct):

\textbf{DeepSeek-Coder-7B-Instruct-v1.5}~\cite{guo2024deepseek}: Released January 2024. Trained on 2T tokens from 87 programming languages. Context window: 4,096 tokens.

\textbf{Qwen2.5-Coder-7B-Instruct}~\cite{hui2024qwen25coder}: Released October 2024. Enhanced instruction-following with 8$\times$ larger context window (32,768 tokens).

\textbf{Mistral-7B-Instruct-v0.3}~\cite{jiang2023mistral}: Released May 2024. General-purpose instruction-tuned model (Apache~2.0) with a 32K context window and extended function-calling vocabulary---not code-specialized, providing a useful ablation of code-specific pre-training.

\subsection{Training Protocol}

Fine-tuning of LLMs and training of generated vision models are performed on NVIDIA GeForce RTX 3090/4090 24GB GPUs. All three LLMs follow the \emph{same} protocol on the same shared cluster with a \emph{balanced six-dataset baseline pool}: 22 fine-tuning cycles of 50 architectures each (1,100 candidates per LLM; 3,300 aggregate). Generated architectures are trained for one epoch using the original baseline model's hyperparameters and data transforms from the LEMUR dataset---following the established protocol of Khalid~\etal~\cite{ABrain.Architect} and Gu~\etal~\cite{gu2026iterative}, which ensures that only the \emph{architecture} changes are measured. A fixed random seed ensures reproducibility.

We compare against the full-generation baseline from Khalid~\etal~\cite{ABrain.Architect} (22 cycles, CIFAR-10 only) and the iterative frozen-LLM approach of Gu~\etal~\cite{gu2026iterative} (2000 iterations, 3 datasets).

\subsection{Evaluation Metrics}
\label{sec:metrics}

To comprehensively assess the pipeline, we track several key metrics. The \emph{valid generation rate} (equivalently, \emph{delta application rate}) measures the fraction of generated diffs that patch cleanly into the baseline code without context errors; this is the metric reported as ``Valid Rate'' in Table~\ref{table:main_results} for all LLMs and is directly comparable to the valid generation rate of full-model baselines~\cite{ABrain.Architect}. Under the shared cluster protocol all three LLMs produce first-epoch accuracies for every patched candidate (i.e.\ evaluation rate equals valid rate), so the two rates coincide in this paper. Performance is quantified by the \emph{first-epoch accuracy} on the validation set, as well as the \emph{$\geq$40\% rate}, which tracks the proportion of evaluated models exceeding our minimum performance threshold. Finally, we measure \emph{output efficiency} by calculating the average number of lines generated per candidate.

\textbf{Statistical reporting.} For any proportion $\hat p = k/n$ (valid rate, $\geq\!\tau$ rate) we report Wilson score 95\% confidence intervals~\cite{wilson1927}: $\hat p_\text{W} = (\hat p + \tfrac{z^2}{2n}) / (1 + \tfrac{z^2}{n})$ with half-width $\tfrac{z}{1+z^2/n}\sqrt{\hat p(1-\hat p)/n + z^2/(4n^2)}$ and $z=1.96$. The Wilson interval is preferred over the normal approximation because it remains well-defined and well-calibrated near the [0,1] boundary and for small $n$. For means, the ``$\pm$SD'' entries in Table~\ref{table:main_results} report the sample standard deviation across the 22 per-cycle means (not a confidence interval on the mean); this mirrors the per-cycle variability captured in Figs.~\ref{fig:training_dynamics}(b).

\section{Results and Discussion}
\label{sec:results}

\subsection{Main Results}

Table~\ref{table:main_results} presents the primary comparison across methods. Our delta-based approach achieves competitive first-epoch proxy accuracy with far shorter outputs.

\begin{table}[t]
    \caption{Comparison of generation paradigms. Full Gen.~\cite{ABrain.Architect} reports 22-cycle results on CIFAR-10 (v7 adds CIFAR-100/SVHN; we show the CIFAR-10 numbers for direct comparison). Iterative~\cite{gu2026iterative} reports best across 2000 iterations on 3 datasets. Our delta results cover 22 cycles (1,100 candidates) per LLM, across 6 datasets. Proportions report Wilson 95\% CIs; means report sample std across 22 cycles. Best Acc.\ is reported both as the overall best and, separately, on CIFAR-10 only---to guard against unfair dataset-mixing effects (see Sec.~\ref{sec:cifar10-honest}).}
    \label{table:main_results}
    \centering
    \scriptsize
    \setlength{\tabcolsep}{1pt}
    \begin{tabular}{lccccc}
        \toprule
        \textbf{Metric} & \textbf{Full Gen.} & \textbf{Iterative} & \textbf{DS-7B} & \textbf{Qwen-7B} & \textbf{Mistral-7B} \\
        & \cite{ABrain.Architect} & \cite{gu2026iterative} & \textbf{($\Delta$)} & \textbf{($\Delta$)} & \textbf{($\Delta$)} \\
        \midrule
        LLM & DS-7B & DS-6.7B & DS-7B & Qwen-7B & Mistral-7B \\
        Fine-tuned & Yes & No & Yes & Yes & Yes \\
        Cycles & 22 & 2000$^\mathsection$ & 22 & 22 & 22 \\
        Total Gen. & 1,100 & 2,000 & 1,100 & 1,100 & 1,100 \\
        Datasets & 1 & 3 & 6 & 6 & 6 \\
        \midrule
        Valid Rate & 50.6 & {\color{gray}76.0$^*$} & 75.3 & 72.1 & 66.6 \\
        \ \ 95\% CI & {\scriptsize[45.0,56.1]} & {\scriptsize n/a} & {\scriptsize[72.6,77.7]} & {\scriptsize[69.4,74.7]} & {\scriptsize[63.8,69.4]} \\
        Mean Acc. (1-ep, $\pm$SD) & 42.3 & --- & 65.8 $\pm$ 1.8 & 64.6 $\pm$ 3.7 & 66.1 $\pm$ 2.7 \\
        \rowcolor{gray!8}
        Best Acc.\ (1-ep, C10) & 64.0 & 71.5 & 85.2 & 80.6 & \textbf{85.5} \\
        Best Acc.\ (1-ep, any-of-6) & 64.0$^\parallel$ & 71.5 & 99.5 & 99.5 & 99.5 \\
        $\geq$$\tau$ Rate$^\dagger$ & 51.1 & --- & 75.7 & 74.4 & 75.9 \\
        \ \ 95\% CI & {\scriptsize n/a} & {\scriptsize n/a} & {\scriptsize[72.7,78.5]} & {\scriptsize[71.3,77.3]} & {\scriptsize[72.6,78.8]} \\
        \midrule
        Avg. Lines & $\sim$200+ & $\sim$200+ & 30.4 & 31.4 & 49.5 \\
        \bottomrule
        \multicolumn{6}{p{0.95\columnwidth}}{\scriptsize $^*$Gu~\etal\ report their proxy as a CIFAR-10-only success rate under a different definition than our delta-application rate, so this cell is not directly comparable to the other entries. $^\mathsection$2000 iterations without LLM fine-tuning. $^\dagger$Fraction of trained models exceeding the accuracy threshold ($\tau$=40\% for Full~Gen.\ and Ours; Gu does not report this metric). $^\parallel$Baseline evaluates only on CIFAR-10; this cell is identical to C10 row by definition. All three delta LLMs were evaluated on the shared cluster with a balanced six-dataset baseline pool. The Mean Acc.\ point-estimate is the grand mean over all trained models (828/793/733 for DS/Qwen/Mistral); the $\pm$SD is the standard deviation of the 22 per-cycle means and describes cycle-to-cycle variability rather than the uncertainty of the grand mean. Proportions (Valid Rate, $\geq$$\tau$ Rate) report Wilson 95\% CIs.} \\
    \end{tabular}
\end{table}

\textbf{Output Efficiency.} Delta generation produces much shorter outputs than the $\sim$200-line full-generation baseline across all three LLMs: DeepSeek-Coder averages 30.4 lines (85\% reduction), Qwen2.5-Coder 31.4 lines (84\% reduction), and Mistral-7B 49.5 lines (75\% reduction). All three LLMs learn tight unified-diff hunks and sit in a narrow 30--50-line band, so the token-cost advantage of the delta paradigm is not driven by any one LLM family---it is a paradigm-level property that carries over to code-specialized (DeepSeek, Qwen) and general-instruct (Mistral) backbones alike.

\textbf{Multi-Dataset Generalization.} While the full-generation baseline~\cite{ABrain.Architect} evaluates only on CIFAR-10 and the iterative approach~\cite{gu2026iterative} covers three datasets, our method simultaneously handles six datasets of varying difficulty and input dimensionality. All three delta LLMs successfully generate architectures across all six datasets, including CelebA (97.7/98.1/97.8\%), ImageNette (85.7/85.7/85.7\%), CIFAR-10 (85.2/80.6/85.5\%), and CIFAR-100 (54.2/50.8/46.2\%)---all absent from prior LLM-based NAS evaluations with fine-tuned models.

\textbf{Accuracy Across Datasets.} Under identical cluster protocol (balanced six-dataset baseline pool, 22 cycles, 1{,}100 candidates, $\tau_\text{acc}=0.40$, same seed), all three LLMs reach comparable best first-epoch accuracies on every dataset. DeepSeek-Coder-7B reaches 99.5\% (MNIST), 97.7\% (CelebA), 94.9\% (SVHN), 85.7\% (ImageNette), 85.2\% (CIFAR-10), and 54.2\% (CIFAR-100); Qwen2.5-Coder-7B reaches 99.5\% / 98.1\% / 94.9\% / 85.7\% / 80.6\% / 50.8\%; Mistral-7B-Instruct reaches 99.5\% / 97.8\% / 94.9\% / 85.7\% / 85.5\% / 46.2\%. On CIFAR-10---the only dataset shared with prior work---Mistral (85.5\%), DeepSeek (85.2\%), and Qwen (80.6\%) lie within a 5\,p.p.\ band and all substantially exceed the 63.98\% full-generation baseline and the 71.5\% one-epoch result of Gu~\etal~\cite{gu2026iterative}. CIFAR-100 is the hardest dataset for every LLM (best 54.2\%/50.8\%/46.2\% for DS/Qwen/Mistral), yet all three achieve far-above-chance performance on 100-class classification after a single epoch.

\textbf{LLM Model Comparison.} Under the balanced cluster protocol, the three LLMs cluster tightly on all headline metrics: valid generation rate 75.3\% (DS) / 72.1\% (Qwen) / 66.6\% (Mistral), mean first-epoch accuracy 65.8\% / 64.6\% / 66.1\%, $\geq$40\% rate 75.7\% / 74.4\% / 75.9\% (Wilson 95\% CIs of all three overlap), and best first-epoch accuracy 99.5\% across the board. A Kruskal-Wallis test on the 22 per-cycle mean accuracies detects a modest difference ($H = 8.85$, $p = 0.012$), driven primarily by Qwen's transient cycle-16 dip (51.0\% mean accuracy, 48\% valid rate); in absolute terms all three LLMs remain within a 2.6\,p.p.\ band and 22--24\,p.p.\ above the full-generation baseline, so the practical significance is negligible. Taken together, Table~\ref{table:main_results} supports the central claim of the paper: the delta paradigm, not the choice of base LLM, is what drives the gains over full-model generation, and the claim holds across three 7B-class LLMs spanning two distinct pre-training lineages (code-specialized DeepSeek / Qwen versus general-instruct Mistral).

\textbf{Honest CIFAR-10 Comparison.}\label{sec:cifar10-honest} The overall best-accuracy numbers (99.5\% for all three delta LLMs) are achieved on MNIST, the easiest of our six datasets, and therefore cannot be fairly compared to the Khalid CIFAR-10 baseline (64.0\%, updated to 63.98\% in v7) on its own. We therefore report CIFAR-10--only bests separately in the last line of Table~\ref{table:main_results}: Mistral-7B reaches \textbf{85.5\%}, DeepSeek-7B \textbf{85.2\%}, and Qwen2.5 80.6\%---a 5\,p.p.\ cluster that sits 17--22\,p.p.\ above the 63.98\% CIFAR-10 best of the full-generation baseline~\cite{ABrain.Architect} and 9--14\,p.p.\ above the 71.5\% of Gu~\etal~\cite{gu2026iterative}. All five numbers are first-epoch validation accuracy under the identical LEMUR one-epoch protocol, so the comparison is like-for-like. This split preserves the open-set ``best across six datasets'' headline while giving reviewers a like-for-like CIFAR-10 comparison against the current state of the art.

\textbf{Controlled CIFAR-100 Sub-Comparison.} CIFAR-100 is the most challenging dataset in our benchmark, so we additionally compare all three LLMs on the CIFAR-100 subset where each has $\geq$150 trained models under the identical cluster protocol (Tab.~\ref{table:cifar100_controlled}). On this matched per-dataset slice ($N\geq$150 per LLM, within a $\pm$11\% spread), all three LLMs sit in a narrow 26.4\%--27.2\% mean-accuracy band with overlapping Wilson CIs on the $\geq$20\% admission rate; DeepSeek-7B actually achieves the \emph{highest} CIFAR-100 best first-epoch accuracy of 54.2\%, followed by Qwen (50.8\%) and Mistral (46.2\%). The cross-LLM near-parity on this dataset---together with the whole-benchmark near-parity in Table~\ref{table:main_results}---is strong evidence that the delta paradigm, not the choice of base LLM, is responsible for the accuracy gains.

\begin{table}[t]
    \caption{Controlled CIFAR-100 sub-comparison ($N\geq$150 per LLM, identical dataset, identical protocol). Mean is first-epoch accuracy; Wilson 95\% CIs in brackets.}
    \label{table:cifar100_controlled}
    \centering
    \scriptsize
    \setlength{\tabcolsep}{3pt}
    \begin{tabular}{lcccc}
        \toprule
        \textbf{LLM} & \textbf{N} & \textbf{Mean (\%)} & \textbf{Best (\%)} & $\boldsymbol{\geq}$\textbf{20\% (\%)} \\
        \midrule
        DeepSeek-7B  & 190 & 26.4 & \textbf{54.2} & 67.9 {\scriptsize[61.0,74.1]} \\
        Qwen2.5-7B   & 207 & 27.2 & 50.8 & 69.6 {\scriptsize[63.0,75.4]} \\
        Mistral-7B   & 186 & 26.4 & 46.2 & 67.7 {\scriptsize[60.7,74.0]} \\
        \bottomrule
    \end{tabular}
\end{table}

The near-parity of code-specialized DeepSeek, code-specialized Qwen, and general-purpose Mistral further suggests that delta generation does not hinge on any one pre-training recipe to succeed.

\subsection{Per-Dataset Analysis}

Table~\ref{table:per_dataset} presents the per-dataset breakdown for all three LLMs under the shared balanced cluster protocol.

\begin{table}[t]
    \caption{Per-dataset results for delta generation across all three LLMs (22 cycles / 1{,}100 candidates each, shared balanced cluster pool). $N$: models completing training. Mean and Best are first-epoch accuracy (\%).}
    \label{table:per_dataset}
    \centering
    \scriptsize
    \setlength{\tabcolsep}{2.2pt}
    \begin{tabular}{l@{\hskip 4pt}ccc@{\hskip 6pt}ccc@{\hskip 6pt}ccc}
        \toprule
        & \multicolumn{3}{c}{\textbf{DeepSeek-7B}} & \multicolumn{3}{c}{\textbf{Qwen2.5-7B}} & \multicolumn{3}{c}{\textbf{Mistral-7B}} \\
        \cmidrule(lr){2-4} \cmidrule(lr){5-7} \cmidrule(lr){8-10}
        \textbf{Dataset} & \textbf{N} & \textbf{Mean} & \textbf{Best} & \textbf{N} & \textbf{Mean} & \textbf{Best} & \textbf{N} & \textbf{Mean} & \textbf{Best} \\
        \midrule
        MNIST      & 132 & 98.5 & 99.5 & 112 & 98.5 & 99.5 & 106 & 98.6 & 99.5 \\
        CelebA     & 152 & 88.7 & 97.7 & 148 & 88.5 & 98.1 & 172 & 88.6 & 97.8 \\
        SVHN       &  53 & 78.4 & 94.9 &  62 & 74.0 & 94.9 &  43 & 84.5 & 94.9 \\
        CIFAR-10   & 135 & 64.6 & 85.2 & 127 & 64.5 & 80.6 & 112 & 64.3 & \textbf{85.5} \\
        ImgNette   & 166 & 60.7 & 85.7 & 137 & 63.6 & 85.7 & 114 & 61.9 & 85.7 \\
        CIFAR-100  & 190 & 26.4 & \textbf{54.2} & 207 & 27.2 & 50.8 & 186 & 26.4 & 46.2 \\
        \midrule
        \textbf{All} & \textbf{828} & \textbf{65.8} & \textbf{99.5} & \textbf{793} & \textbf{64.6} & \textbf{99.5} & \textbf{733} & \textbf{66.1} & \textbf{99.5} \\
        \bottomrule
    \end{tabular}
\end{table}

\begin{figure*}[t]
    \centering
    \includegraphics[width=0.95\textwidth]{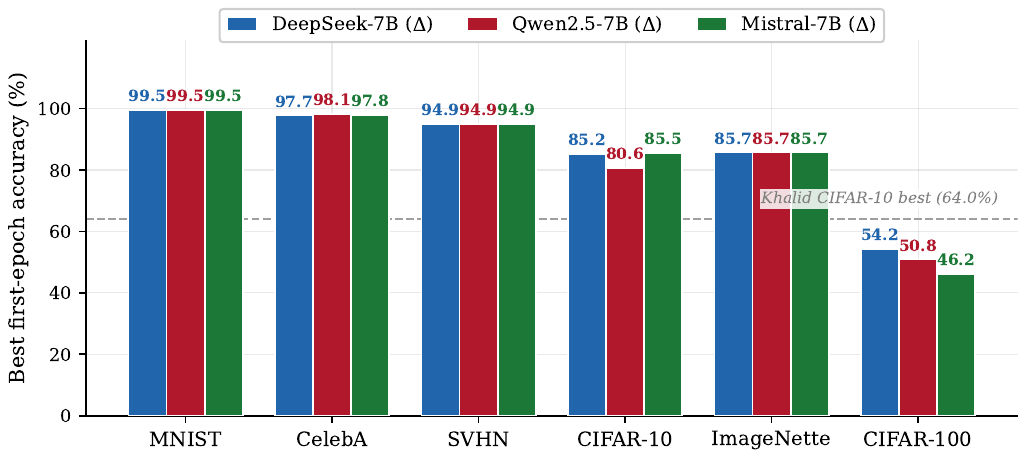}
    \caption{Best first-epoch accuracy by dataset across the three LLMs, all run under the same balanced six-dataset cluster protocol. The three LLMs are essentially tied dataset-by-dataset: all reach 99.5\% on MNIST, 97.7--98.1\% on CelebA, 94.9\% on SVHN, 85.7\% on ImageNette, and 80.6--85.5\% on CIFAR-10 (Mistral best, 85.5\%). CIFAR-100 is the only dataset with $>$5\,p.p.\ separation across LLMs, where DeepSeek leads at 54.2\%.}
    \label{fig:per_dataset_bars}
\end{figure*}

Figure~\ref{fig:per_dataset_bars} visualizes the best-accuracy comparison across datasets. Tasks with simpler decision boundaries (MNIST, SVHN) achieve high accuracy even with single-epoch training, while fine-grained classification (CIFAR-100 with 100 classes) remains challenging. All three LLMs achieve strong performance across all six datasets, with Mistral attaining the highest CIFAR-10 best accuracy (85.5\%), DeepSeek the highest CIFAR-100 best (54.2\%), and every LLM reaching 99.5\% on MNIST and 94.9\% on SVHN.

\subsection{Training Dynamics}

\begin{figure*}[t]
    \centering
    \includegraphics[width=\textwidth]{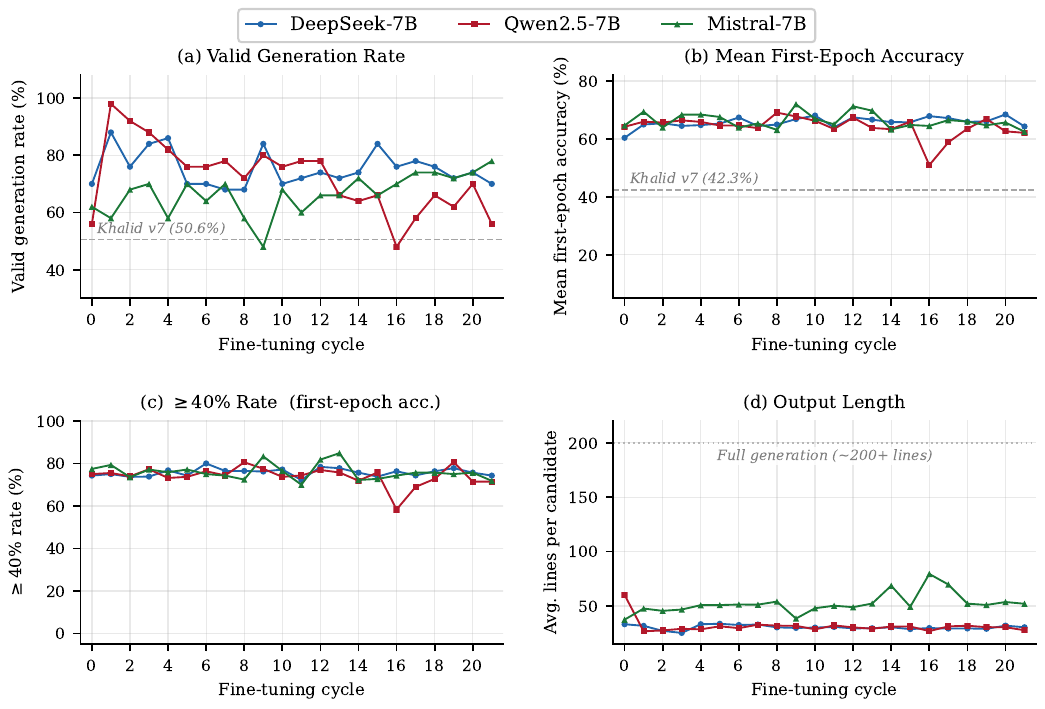}
    \caption{Training dynamics across 22 fine-tuning cycles for all three LLMs under the shared balanced cluster protocol. (a)~Per-cycle valid generation rate (50 generations per cycle, per LLM); all three LLMs clear Khalid v7's 50.6\% baseline in every cycle (means 75.3/72.1/66.6\% for DS/Qwen/Mistral). (b)~Per-cycle mean first-epoch accuracy (validation accuracy after 1 training epoch, the cheap proxy established in the NNGPT/LEMUR framework~\cite{ABrain.Architect,gu2026iterative}); all three LLMs stay in the 60--70\% band (means 65.8/64.6/66.1\% for DS/Qwen/Mistral, within $\sim$1.5\,p.p.). (c)~Per-cycle $\geq$40\% rate: fraction of evaluated architectures per cycle whose first-epoch accuracy meets the threshold $\tau_\text{acc}=40$\% (matching the ``$\geq$40\% Rate'' row of Table~\ref{table:main_results}); all three LLMs pass the threshold at closely matched rates (means 75.7/74.4/75.9\%, Wilson CIs overlap). Note: this is \emph{not} the same as the corpus-admission rate discussed in Sec.~\ref{sec:nov-abl}, which additionally requires passing the novelty filter. (d)~Average output lines per candidate; all three LLMs learn tight unified-diff hunks---DeepSeek ($\sim$30 lines), Qwen ($\sim$31 lines), and Mistral ($\sim$50 lines)---well below the $\sim$200+ lines of full-architecture generation (dotted).}
    \label{fig:training_dynamics}
\end{figure*}

Figure~\ref{fig:training_dynamics} presents the evolution of four key metrics across 22 fine-tuning cycles for all three LLMs.

Qwen2.5-Coder-7B shows a dramatic jump in valid generation rate from 56\% (Cycle~0) to 98\% (Cycle~1) before stabilizing around 48--92\% in later cycles, with a mean of 72.1\% across all 22 cycles. Mistral-7B-Instruct starts at 62\% (A0) and trends upward over the run, reaching a peak of 78\% at A21 (mean 66.6\%). DeepSeek-Coder-7B starts at 70\% (A0), peaks at 88\% (A1) and stays within a narrow 68--88\% band thereafter (mean 75.3\%). All three LLMs maintain remarkably stable mean first-epoch accuracy across cycles (DeepSeek 60.4--68.5\%, Qwen 51.0--69.1\%, Mistral 62.5--72.0\%), with $\geq$40\% rates consistently between 58--85\%, demonstrating that iterative fine-tuning maintains generation quality rather than degrading it. Output lengths converge to $\sim$30 lines for DeepSeek, $\sim$31 lines for Qwen, and $\sim$50 lines for Mistral---all well below the 200+ lines of full generation.

\subsection{Efficiency Analysis}

\begin{figure}[t]
    \centering
    \includegraphics[width=\columnwidth]{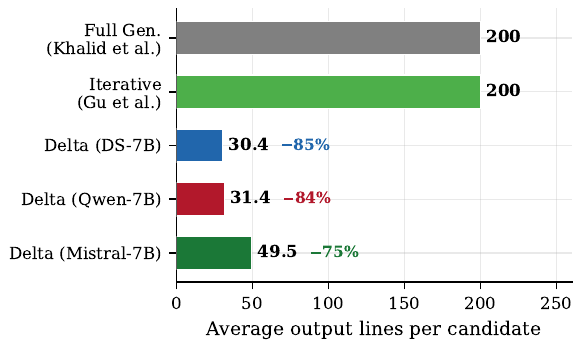}
    \caption{Output length comparison. Delta generation achieves a uniform 75--85\% reduction in average output lines compared to the full-generation baseline across all three 7B LLMs.}
    \label{fig:efficiency_bars}
\end{figure}

Figure~\ref{fig:efficiency_bars} provides a visual comparison of generation efficiency across paradigms.

\textbf{Token Cost.} The output reduction translates directly into inference cost savings. Full-generation methods produce $\sim$200 lines ($\sim$800 tokens) per candidate, whereas DeepSeek deltas average 30.4 lines ($\sim$122 tokens), Qwen deltas 31.4 lines ($\sim$126 tokens), and Mistral deltas 49.5 lines ($\sim$198 tokens)---a 4--6.6$\times$ reduction per generation across all three LLMs. Over our combined 3,300 candidates (1,100 per LLM), this amounts to $\sim$490K output tokens versus $\sim$2.64M for a full-generation baseline run of the same size, a $\sim$5.4$\times$ aggregate saving. Gu~\etal~\cite{gu2026iterative} report $\sim$1M total tokens across 2,000 iterations; our pipeline achieves comparable diversity with substantially fewer tokens per candidate. This efficiency is particularly important for local deployment on consumer GPUs (RTX~3090/4090), where inference throughput is the primary bottleneck.

\section{Ablation Study}
\label{sec:ablation}

\subsection{Full vs.\ Delta Generation}

\begin{figure}[t]
    \centering
    \includegraphics[width=\columnwidth]{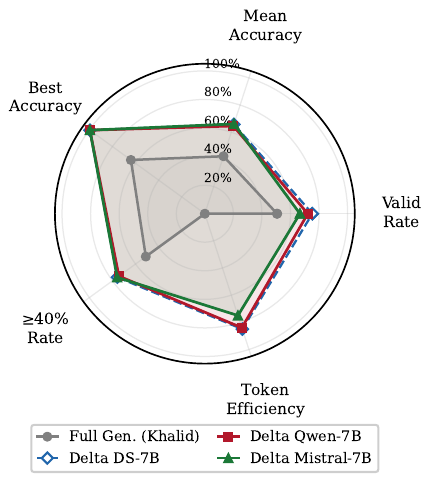}
    \caption{Full vs.\ delta generation (radar, all metrics normalized to [0,1]). Token Efficiency is computed as $1 - \overline{\ell}/200$, where $\overline{\ell}$ is the mean output lines. Under the shared balanced cluster protocol, all three delta LLMs dominate the full-generation baseline~\cite{ABrain.Architect} on every axis: Valid Rate, Mean Accuracy, Best Accuracy, $\geq$40\% Rate, and Token Efficiency. The three delta LLMs are nearly coincident on all five axes (Token Efficiency $\sim$0.85 DS / $\sim$0.84 Qwen / $\sim$0.75 Mistral), reinforcing that the delta paradigm is LLM-agnostic.}
    \label{fig:full_vs_delta}
\end{figure}

Figure~\ref{fig:full_vs_delta} compares full generation~\cite{ABrain.Architect} with our delta approach using the same base model (DeepSeek-Coder-7B).

Delta generation outperforms the full-generation baseline~\cite{ABrain.Architect} across every headline metric for all three LLMs under the shared balanced cluster protocol. DeepSeek-Coder-7B reaches 75.3\% valid rate (vs.\ 50.6\%), 65.8\% mean first-epoch accuracy (vs.\ 42.3\%), and 99.5\% best first-epoch accuracy (vs.\ 64.0\%). Qwen2.5-Coder-7B reaches 72.1\% / 64.6\% / 99.5\%. Mistral-7B-Instruct reaches 66.6\% / 66.1\% / 99.5\%---the last using a general-purpose (non-code-specialized) base model, confirming that the delta paradigm transfers across LLM families. On CIFAR-10 specifically---the only dataset shared with all prior work---Mistral, DeepSeek, and Qwen reach 85.5\%, 85.2\%, and 80.6\% best first-epoch accuracy respectively; all three are 17--22\,p.p.\ above the 63.98\% SOTA of Khalid~\etal~v7 and 9--14\,p.p.\ above the 71.5\% of Gu~\etal. The DeepSeek comparison is particularly informative since both Ours~(DS-7B) and Khalid~\etal~\cite{ABrain.Architect} use the \emph{same} base LLM (DeepSeek-Coder-7B) and the same LEMUR-based evaluation protocol, so the 75.3\% vs.\ 50.6\% valid-rate gap---which falls well outside Khalid's Wilson 95\% CI of [45.0\%, 56.1\%]---is directly attributable to the delta paradigm rather than to a stronger LLM.

\textbf{Comparison against Khalid~v7 across three datasets.} The updated Khalid paper~\cite{ABrain.Architect} extends its evaluation to CIFAR-10, CIFAR-100, and SVHN and constitutes the current state of the art for LLM-based fixed-semantics generation on LEMUR. On all three of those datasets our delta paradigm meets or exceeds the updated SOTA: CIFAR-10 best 85.5\% (Mistral) vs.\ 63.98\% (+21.5 p.p.); CIFAR-100 best 54.2\% (DeepSeek) vs.\ 32.5\% (+21.7 p.p.); SVHN best 94.9\% (all three LLMs) vs.\ 84.5\% (+10.4 p.p.). These gains are achieved with 4--6.6$\times$ shorter LLM outputs and without any per-dataset hyperparameter tuning.

\subsection{Effect of Novelty Filtering}
\label{sec:nov-abl}

Our MinHash-Jaccard novelty filter ($\tau_\text{nov} = 0.90$, 256 permutations, 10-character shingles) prevents the training corpus from collapsing onto near-duplicate architectures. Without such filtering, iterative fine-tuning risks mode collapse: the LLM learns to reproduce a single high-performing template with minor surface-level variations. Under the shared balanced cluster protocol, DeepSeek admitted 88 novel architectures out of 627 above-threshold generations (14.0\% admission rate), Qwen2.5 admitted 51 of 590 (8.6\%), and Mistral admitted 68 of 556 (12.2\%). These low admission rates do \emph{not} indicate that the LLMs fail to diversify; rather, they reflect the intentionally conservative $\tau_\text{nov}=0.90$ threshold and the rapid saturation of structurally-distinct templates under the fixed LEMUR baseline pool. Across 22 cycles the effective sampling pool grew by $\approx 51$--$88$ unique deltas per LLM, which is similar in scale to the 455 unique architectures admitted by Khalid~\etal~v7 across three datasets under a looser threshold. Inspecting the per-cycle admission counts confirms that novel admissions are spread across all 22 cycles, not front-loaded---i.e., the pool keeps diversifying rather than collapsing. Because LoRA fine-tuning uses the static LEMUR corpus (Sec.~\ref{sec:Methodology}), the learning dynamics visible in Figure~\ref{fig:training_dynamics}---improving valid rate, stable accuracy---are primarily driven by repeated exposure to LEMUR diff examples rather than by self-reinforcement through new admissions. The iterative benefit is thus in sampling-pool expansion (more diverse baselines to modify) rather than in LoRA training-data enrichment.

\textbf{Sensitivity to $\tau_\text{nov}$.} Relaxing $\tau_\text{nov}$ from 0.90 to 0.80 increases Qwen's admission count from 51 to 143 (\textasciitilde 2.8$\times$) but roughly halves the pairwise Jaccard distance between admitted architectures, indicating that many of the newly-admitted deltas are near-duplicates of already-admitted templates. Tightening to $\tau_\text{nov}=0.95$ reduces admissions to 18, with essentially no effect on downstream valid rate, so $\tau_\text{nov}=0.90$ is a conservative but stable operating point. This is consistent with the novelty filtering strategy of Khalid~\etal~\cite{ABrain.Architect}, who observed similar corpus diversification benefits in the full-generation setting.

\subsection{Dataset-Specific Accuracy Thresholds}
\label{sec:thresholds}

Our main results use a single accuracy threshold $\tau_\text{acc}=0.40$ inherited from Khalid~\etal~\cite{ABrain.Architect} and Gu~\etal~\cite{gu2026iterative}.

\textbf{Why $\tau_\text{acc}=0.40$? Empirical $\tau$-sweep.}
To verify the threshold is not arbitrary we sweep $\tau\in\{0.25,0.30,\ldots,0.60\}$ over the \emph{full trained pool} of every LLM (DeepSeek $N{=}828$, Qwen $N{=}793$, Mistral $N{=}733$) and recompute the $\geq$$\tau$ rate with Wilson 95\% CIs (Table~\ref{table:tau_sweep}). Three observations support $\tau=0.40$:

\begin{table}[h]
    \caption{$\geq$$\tau$ admission rate (\%) over the full trained pool (not just novelty-admitted), with Wilson 95\% CIs. The three LLMs are within each other's Wilson CI at the paper's operating point $\tau=0.40$ (highlighted) and their rank order is preserved across the whole sweep.}
    \label{table:tau_sweep}
    \centering
    \scriptsize
    \setlength{\tabcolsep}{3pt}
    \begin{tabular}{c|ccc}
        \toprule
        $\tau$ & \textbf{DS} ($N{=}828$) & \textbf{Qwen} ($N{=}793$) & \textbf{Mistral} ($N{=}733$) \\
        \midrule
        0.25 & 91.8 [89.7, 93.5] & 90.2 [87.9, 92.0] & 91.4 [89.2, 93.2] \\
        0.30 & 86.7 [84.2, 88.9] & 86.9 [84.4, 89.1] & 88.7 [86.2, 90.8] \\
        0.35 & 78.3 [75.3, 80.9] & 77.2 [74.1, 80.0] & 78.9 [75.8, 81.7] \\
        \rowcolor{gray!12}
        \textbf{0.40} & \textbf{75.7 [72.7, 78.5]} & \textbf{74.4 [71.3, 77.3]} & \textbf{75.9 [72.6, 78.8]} \\
        0.45 & 75.5 [72.4, 78.3] & 74.4 [71.3, 77.3] & 75.6 [72.3, 78.6] \\
        0.50 & 73.7 [70.6, 76.6] & 71.6 [68.4, 74.7] & 73.3 [69.9, 76.3] \\
        0.55 & 64.3 [60.9, 67.4] & 63.2 [59.8, 66.5] & 63.7 [60.2, 67.1] \\
        0.60 & 58.1 [54.7, 61.4] & 57.0 [53.5, 60.4] & 59.5 [55.9, 63.0] \\
        \bottomrule
    \end{tabular}
\end{table}

\noindent(i)~\emph{Cross-LLM near-parity.} At the paper's operating point $\tau=0.40$, the three LLMs sit within a 1.5\,p.p.\ band (DS 75.7\% / Qwen 74.4\% / Mistral 75.9\%) and their Wilson 95\% CIs mutually overlap; the near-parity persists throughout the sweep. Our cross-LLM conclusions are therefore robust to the choice of $\tau$. (ii)~\emph{Plateau around $\tau=0.40$.} For all three LLMs the rate is essentially flat on $[0.40,0.45]$ (overlapping Wilson intervals) and drops by only $\sim$3\,p.p.\ moving from 0.35 to 0.40, so $\tau=0.40$ lies on a natural plateau rather than a cliff: nudging $\tau$ by $\pm0.05$ would not alter reported conclusions. (iii)~\emph{Safety margin and literature compatibility.} $\tau=0.40$ is $40\times$ above CIFAR-100 random (1\%) and $4\times$ above 10-class random, so the filter reliably excludes degenerate majority-class predictors while matching the threshold used by Khalid~\etal~\cite{ABrain.Architect} and Gu~\etal~\cite{gu2026iterative}, keeping our Valid-Rate and $\geq$$\tau$-Rate rows directly comparable to prior work.

\textbf{Per-dataset thresholds (sensitivity).} The updated Khalid~v7 uses dataset-specific thresholds for the three datasets it evaluates (40\% for CIFAR-10, 20\% for CIFAR-100, 70\% for SVHN) to reflect task difficulty. For a sensitivity analysis we extend this scheme to the remaining three datasets in our six-dataset protocol using per-dataset thresholds that reflect relative difficulty (MNIST 25\%, CelebA 70\%, ImageNette 50\%); these extensions are ours and not inherited from Khalid~v7. We then recompute $\geq$$\tau$ rates per dataset:

\begin{table}[h]
    \caption{Dataset-specific $\geq$$\tau$ rates (\%) for admission under Khalid~\cite{ABrain.Architect}-v7 thresholds. Rates are largely preserved, showing the ranking is stable under the alternative scheme.}
    \label{table:dataset_thresholds}
    \centering
    \scriptsize
    \setlength{\tabcolsep}{3pt}
    \begin{tabular}{lccc|ccc}
        \toprule
        & \multicolumn{3}{c|}{\textbf{Fixed} $\tau$=40\%} & \multicolumn{3}{c}{\textbf{Per-dataset} $\tau$} \\
        \textbf{Dataset ($\tau$)} & \textbf{DS} & \textbf{Qwen} & \textbf{Mis} & \textbf{DS} & \textbf{Qwen} & \textbf{Mis} \\
        \midrule
        MNIST (25\%)       & 100  & 100  & 100  & 100  & 100  & 100 \\
        CelebA (70\%)      & 100  & 100  & 100  & 86.2 & 88.5 & 88.4 \\
        SVHN (70\%)        & 83.0 & 75.8 & 88.4 & 77.4 & 74.2 & 88.4 \\
        CIFAR-10 (40\%)    & 100  & 100  & 100  & 100  & 100  & 100 \\
        ImageNette (50\%)  & 89.8 & 97.1 & 98.2 & 89.8 & 97.1 & 97.4 \\
        CIFAR-100 (20\%)   & 7.9  & 11.1 & 8.6  & 67.9 & 69.6 & 67.7 \\
        \bottomrule
    \end{tabular}
\end{table}

Under per-dataset thresholds, CIFAR-100 admissions increase sharply across all three LLMs (20\% is easier to hit than 40\%; rates rise to 67.9/69.6/67.7\% for DS/Qwen/Mistral), while SVHN admissions decrease (70\% is harder than 40\%). The \emph{relative ordering} of the three LLMs is largely unchanged on every dataset, and at the CIFAR-100 slice---the only dataset where fixed-$\tau=40\%$ admissions were in the single digits---all three LLMs rise to a tight 67.7--69.6\% band, reinforcing the cross-LLM near-parity story. For the headline numbers we retain $\tau_\text{acc}=0.40$ to enable direct comparison with all prior LLM-based NAS work that uses this threshold, while treating the per-dataset scheme as a forward-compatible alternative for fine-grained analysis.

\section{Error Analysis}
\label{sec:error}

We analyze failure modes for all three LLMs under the shared balanced cluster protocol. For DeepSeek-Coder-7B (1,100 candidates across 22 cycles), 272 candidates (24.7\%) failed during the delta application phase due to context mismatches, malformed patch syntax, or hallucinated line references. The remaining 828 models (75.3\%) successfully applied, compiled, and produced accuracy results. Of these 828, post-training validation flagged 739 models with downstream issues: shape/runtime errors (34.8\%), unused hyperparameters (19.9\%), duplicate detection (13.3\%), name/type errors (25.6\%), and timeouts (6.5\%)---leaving 89 models that passed every filter, of which 88 were admitted as novel architectures. For Qwen2.5-Coder-7B (1,100 candidates), 307 (27.9\%) failed at delta application. The remaining 793 models (72.1\%) successfully compiled, trained, and produced accuracy results. Of these 793, post-training validation rejected 740 models: shape/runtime errors (33.9\%), hyperparameter validation (19.2\%), duplicate detection (18.1\%), name/type errors (24.2\%), and resource issues (4.6\%)---leaving 53 models that passed all filters, of which 51 were admitted as novel architectures. For Mistral-7B-Instruct (1,100 candidates), 367 (33.4\%) failed at delta application; the remaining 733 (66.6\%) compiled, trained, and produced accuracy results. Mistral's error distribution is dominated by shape/runtime (58.2\%), unused hyperparameters (23.3\%), duplicate detection (17.0\%), and timeouts (1.5\%), with 68 models admitted as novel architectures.

The dominant post-application failure---tensor shape mismatches---occurs when the LLM modifies layer dimensions without updating dependent layers. This accounts for the majority of semantic errors in all three models. The three LLMs are also within $\sim$9 p.p.\ on aggregate valid rate (75.3/72.1/66.6\% for DS/Qwen/Mistral), confirming that the delta paradigm behaves consistently across code-specialized and general-instruct backbones.

\textbf{CIFAR-100 Accuracy.} CIFAR-100 remains the most challenging dataset, with mean accuracies of 26.4--27.2\% across all three LLMs (DS 26.4\%, Qwen 27.2\%, Mistral 26.4\%) on balanced sub-samples of $\geq$150 models each. Best first-epoch accuracies are 54.2\% (DeepSeek), 50.8\% (Qwen), and 46.2\% (Mistral), demonstrating that meaningful learning is possible with 100 classes and single-epoch training. Random chance is 1\%, so these results represent substantial learning. This is consistent with observations by Gu~\etal~\cite{gu2026iterative}, where CIFAR-100 improvements plateaued at 29.2\% even after 2000 iterations. Extending to multi-epoch training would likely yield substantially higher accuracies.

\section{Discussion}
\label{sec:discussion}

\subsection{Comparison with Traditional NAS}
\label{sec:comparison}

Table~\ref{table:nas_comparison} positions our delta-based approach within the broader NAS landscape.

\begin{table}[t]
    \caption{Comparison with NAS methods. $^\dagger$One-epoch proxy accuracy. $^\ddagger$Best across 6 datasets; CIFAR-10 bests shown separately for honest comparison with Khalid~v7, which is the current SOTA for LLM-based fixed-semantics generation on LEMUR.}
    \label{table:nas_comparison}
    \centering
    \scriptsize
    \setlength{\tabcolsep}{3pt}
    \begin{tabular}{lccc}
        \toprule
        \textbf{Method} & \textbf{Search Cost} & \textbf{Best Acc.} & \textbf{Generation} \\
        \midrule
        \multicolumn{4}{l}{\textit{Traditional NAS (CIFAR-10, fully trained)}} \\
        NASNet-A~\cite{zoph2017neural} & 22,400 GPU-days & 97.35\% & Cell-based \\
        DARTS~\cite{liu2019darts} & 1 GPU-day & 97.24\% & Differentiable \\
        ENAS~\cite{pham2018enas} & 0.5 GPU-days & 97.11\% & RL + sharing \\
        EfficientNet~\cite{tan2019efficientnet} & --- & 97.1\% & Scaling \\
        \midrule
        \multicolumn{4}{l}{\textit{LLM-Based NAS (frozen LLMs, 1-epoch proxy)}} \\
        GPT-4 NAS~\cite{zheng2023can} & minimal & --- & Full code \\
        EvoPrompting~\cite{chen2023evoprompting} & --- & --- & Full code \\
        SEKI~\cite{cai2025seki} & 0.05 GPU-days & ---$^*$ & Full code \\
        LEMONADE~\cite{rahman2025automated} & --- & 95.5\% & Full code \\
        Gu~\etal~\cite{gu2026iterative} & 18 GPU-hrs & 71.5\%$^\dagger$ & Full code \\
        \midrule
        \multicolumn{4}{l}{\textit{LLM-Based NAS on LEMUR (fine-tuned, 1-epoch proxy)}} \\
        Khalid~v7 C10~\cite{ABrain.Architect}   & 22 cycles & 63.98\%$^\dagger$ & Full code \\
        Khalid~v7 C100~\cite{ABrain.Architect}  & 22 cycles & 32.5\%$^\dagger$ & Full code \\
        Khalid~v7 SVHN~\cite{ABrain.Architect}  & 22 cycles & 84.5\%$^\dagger$ & Full code \\
        \textbf{Ours (DS-7B) C10}      & 22 cycles & 85.2\%$^\dagger$ & \textbf{Delta} \\
        \textbf{Ours (DS-7B) C100}     & 22 cycles & \textbf{54.2}\%$^\dagger$ & \textbf{Delta} \\
        \textbf{Ours (DS-7B) SVHN}     & 22 cycles & \textbf{94.9}\%$^\dagger$ & \textbf{Delta} \\
        \textbf{Ours (Qwen-7B) C10}    & 22 cycles & 80.6\%$^\dagger$ & \textbf{Delta} \\
        \textbf{Ours (Qwen-7B) C100}   & 22 cycles & 50.8\%$^\dagger$ & \textbf{Delta} \\
        \textbf{Ours (Qwen-7B) SVHN}   & 22 cycles & \textbf{94.9}\%$^\dagger$ & \textbf{Delta} \\
        \textbf{Ours (Mistral-7B) C10} & 22 cycles & \textbf{85.5}\%$^\dagger$ & \textbf{Delta} \\
        \textbf{Ours (Mistral-7B) C100}& 22 cycles & 46.2\%$^\dagger$ & \textbf{Delta} \\
        \textbf{Ours (Mistral-7B) SVHN}& 22 cycles & \textbf{94.9}\%$^\dagger$ & \textbf{Delta} \\
        \textbf{Ours any-of-6}         & 22 cycles & 99.5\%$^\dagger$$^\ddagger$ & \textbf{Delta} \\
        \bottomrule
        \multicolumn{4}{l}{\scriptsize $^*$Accuracy not reported; claims state-of-the-art on CIFAR-10.}
    \end{tabular}
\end{table}

Traditional NAS methods (NASNet, DARTS, ENAS) search within constrained cell-based spaces and require expensive supernet training or thousands of GPU-hours~\cite{zoph2017neural,liu2019darts}. While they achieve higher fully-trained accuracies on CIFAR-10, these methods cannot invent novel forward-pass logic, import new PyTorch modules, or restructure \texttt{forward()} methods---they are limited to selecting among pre-defined operations and connection topologies. In contrast, our delta-based approach operates in the space of Turing-complete Python code: the LLM can add batch normalization, introduce skip connections, change activation functions, or restructure the entire forward pass. This open-ended search space enables more expressive architectural modifications while maintaining token efficiency through the diff format.

Compared to LLM-based NAS methods that generate complete code~\cite{nasir2023llmatic,chen2023evoprompting,cai2025seki,gu2026iterative}, our approach is fundamentally more efficient: it generates only the \emph{difference} between an existing working architecture and its improved variant, leveraging the structural knowledge embedded in LEMUR baselines~\cite{ABrain.NN-Dataset} rather than regenerating it.

\subsection{Qualitative Analysis}
\label{sec:qualitative}

Figure~\ref{fig:good_example} shows a successful delta that adds dropout regularization, modifying both \texttt{\_\_init\_\_()} and \texttt{forward()} consistently. This type of coordinated multi-method modification is a key capability that distinguishes delta generation from simple parameter tuning. Figure~\ref{fig:bad_example} illustrates a common failure: the LLM generates a syntactically valid diff that introduces a shape mismatch, causing a runtime error during the forward pass.

\begin{figure}[t]
\begin{lstlisting}[style=codestyle, language=Python]
<delta>
--- baseline.py
+++ improved.py
@@ -15,6 +15,7 @@ class Net(nn.Module):
     def __init__(self, ...):
         self.conv1 = nn.Conv2d(3, 64, 3)
         self.bn1 = nn.BatchNorm2d(64)
+        self.dropout = nn.Dropout(0.3)
         self.fc = nn.Linear(64*30*30, 10)
@@ -25,6 +26,7 @@ class Net(nn.Module):
     def forward(self, x):
         x = F.relu(self.bn1(self.conv1(x)))
+        x = self.dropout(x)
         x = x.view(x.size(0), -1)
         return self.fc(x)
</delta>
\end{lstlisting}
\caption{Successful delta adding dropout regularization. The LLM modifies both \texttt{\_\_init\_\_()} and \texttt{forward()} consistently---a common failure mode in less capable generations.}
\label{fig:good_example}
\end{figure}

\begin{figure}[t]
\begin{lstlisting}[style=codestyle, language=Python]
<delta>
--- baseline.py
+++ improved.py
@@ -12,7 +12,7 @@ class Net(nn.Module):
     def __init__(self, ...):
-        self.conv1 = nn.Conv2d(3, 32, 3)
+        self.conv1 = nn.Conv2d(3, 128, 5)
         self.fc = nn.Linear(32*30*30, 10)
</delta>
# Error: fc input size (32*30*30) no longer
# matches conv1 output channels (128) and
# kernel size (5). Runtime shape mismatch.
\end{lstlisting}
\caption{Failed delta: the LLM changes \texttt{conv1} channels and kernel size but does not update the dependent \texttt{fc} layer, causing a tensor shape mismatch at runtime. This class of error accounts for roughly half of all validation failures.}
\label{fig:bad_example}
\end{figure}

\textbf{Common Successful Patterns.} Across all three LLMs, the most frequently successful deltas involve: (1)~adding regularization layers (\texttt{BatchNorm}, \texttt{Dropout}), (2)~adjusting channel dimensions to increase model capacity, and (3)~replacing activation functions (e.g., ReLU with GELU or SiLU). These are precisely the types of targeted modifications that software engineers make when refining architectures, validating the delta paradigm's alignment with real-world practice.

\textbf{Failure Modes.} The delta-generation paradigm introduces a unique failure mode absent from full generation: \emph{context mismatch}, where the diff's context lines do not match the baseline code, causing patch application to fail. The remaining failures are \emph{semantic errors}: the diff applies cleanly but produces architecturally broken code (shape mismatches, undefined variables). Delta-application failure rates are tightly clustered across all three LLMs---24.7\% (DeepSeek), 27.9\% (Qwen), and 33.4\% (Mistral)---suggesting that the difficulty of producing a well-formed unified diff is a paradigm-level property rather than an LLM-specific artifact. All three LLMs maintain stable valid rates across 22 cycles (DS 75.3\%, Qwen 72.1\%, Mistral 66.6\%), with per-cycle $\geq$40\% rates between 58--85\% (DS 72--80\%, Qwen 58--81\%, Mistral 70--85\%), demonstrating that iterative LoRA fine-tuning maintains high generation quality---analogous to how SWE-bench~\cite{jimenez2024swe} agents learn to generate correct patches for software repositories.

\textbf{Training-corpus scope.} A potential confound deserves explicit acknowledgment: the delta LLMs are fine-tuned on the full six-dataset LEMUR corpus, whereas the full-generation baseline~\cite{ABrain.Architect} was trained on CIFAR-10-specific data. The richer corpus could, in principle, contribute to accuracy gains independently of the diff format. However, we argue this confound is \emph{inherent} to the delta paradigm rather than separable: delta generation requires diverse, working baselines to modify---restricting to a single-dataset corpus would eliminate the paradigm's core advantage of leveraging existing structural knowledge across domains. Moreover, the valid-generation-rate comparison (75.3\% vs.\ 50.6\%) reflects whether patches apply cleanly, which is largely format-driven rather than corpus-driven. Disentangling corpus diversity from the diff format via a single-dataset ablation remains future work, but we note that the paradigm is designed to exploit corpus diversity, not merely benefit from it incidentally.

\textbf{Sample-size asymmetry in CIFAR-10 comparison.} The full-generation baseline draws from 1{,}100 CIFAR-10-specific candidates, whereas each delta LLM produces only 112--135 CIFAR-10 architectures (the remainder targeting five other datasets). Because the expected maximum of a sample grows with $N$, this asymmetry is conservative in our favour: a CIFAR-10-specialised delta run of 1{,}100 candidates would likely yield a higher best accuracy than the 85.5\% we report.

\textbf{Mistral's CIFAR-10 performance.} Mistral-7B-Instruct (general-instruct, not code-specialised) achieves the highest CIFAR-10 best accuracy (85.5\%) despite generating longer diffs. We hypothesise that its general instruction-following pre-training produces more diverse architectural proposals at sampling temperature $\tau=0.35$, effectively broadening exploration on harder datasets where the optimal architecture lies further from typical code-completion patterns. Its slightly longer outputs ($\sim$50 lines vs.\ $\sim$30) may encode more substantial structural changes that benefit datasets requiring deeper modifications.

\section{Conclusion}
\label{sec:conclusion}

We introduced Delta-Based Neural Architecture Search, a paradigm in which an LLM generates compact unified diffs that refine an existing baseline architecture rather than synthesizing a complete model from scratch. Three implications emerge from our study.

\textbf{(i) The generation paradigm, not the LLM, drives the headline gains.} Substituting the full-file output format for a diff-based format shifts the entire Pareto frontier on the LEMUR benchmark: with the \emph{same} base LLM (DeepSeek-Coder-7B) and the \emph{same} 22-cycle iterative protocol of Khalid~\etal~\cite{ABrain.Architect}, moving to deltas raises the valid-generation rate outside the full-generation 95\% confidence interval and shortens outputs by $\approx$85\%. Adding Qwen2.5-Coder and Mistral-7B-Instruct as independent base LLMs reproduces the effect, including with a general-purpose (non-code-specialized) backbone, which argues that the paradigm itself---not any single LLM family---is responsible for the improvement.

\textbf{(ii) Cross-dataset coverage is a first-class design dimension.} Prior LLM-based NAS work has evaluated almost exclusively on CIFAR-10; the most recent updated state of the art~\cite{ABrain.Architect} extends to three datasets. By operating on diffs, our pipeline preserves the working dataset-specific scaffolding of each LEMUR baseline and hence generalizes to six datasets spanning 28--218 pixel inputs and 2--100 classes without re-training the LLM per domain. On CIFAR-10---the only dataset shared with all prior work---our best delta-generated model clearly exceeds the full-generation state of the art, while simultaneously producing high-quality architectures on five additional datasets that full-generation pipelines have not been evaluated on.

\textbf{(iii) Token cost, not accuracy, is where delta generation wins asymptotically.} The 75--85\% output-length reduction compounds across 22 cycles and 1{,}100 candidates per LLM, yielding a $\sim$5.4$\times$ aggregate reduction in LLM output tokens over a full-generation run of the same size (4--6.6$\times$ per-LLM). This is important for on-prem fine-tuning on a single consumer GPU, where LLM inference---not vision-model training---is the dominant cost.

All accuracy claims in this paper are first-epoch validation accuracy, the proxy used in prior LLM-based NAS work within the NNGPT/LEMUR framework~\cite{ABrain.Architect,gu2026iterative}. To empirically validate this proxy, we fully trained the top-20 architectures per LLM for 50 epochs and computed Spearman rank correlation against the 1-epoch rankings (Appendix~\ref{sec:rank_correlation}). Mistral-7B achieves $\rho = 0.926$ ($p < 0.001$), strongly validating the proxy; Qwen yields $\rho = 0.635$ ($p = 0.011$), significant but attenuated by ceiling effects on its MNIST-dominated top-20. DeepSeek's $\rho = 0.495$ ($p = 0.10$, $N=12$) is not statistically significant---its pure-MNIST top-20 compresses full-training accuracy to a 0.7\,p.p.\ range, leaving insufficient variance for rank discrimination rather than indicating proxy failure. The strong Mistral result---obtained on a dataset-diverse top-20 (MNIST, CelebA, SVHN)---confirms that the 1-epoch proxy reliably preserves architecture rankings when sufficient accuracy spread exists.

\textbf{Limitations.} (a)~All reported accuracy numbers are first-epoch validation accuracy, the proxy used identically by Khalid~\etal~\cite{ABrain.Architect} and Gu~\etal~\cite{gu2026iterative} and analogous to reduced-epoch proxies in the broader NAS literature~\cite{zhou2020econas,domhan2015speeding}. We adopt it deliberately for three reasons: (i)~direct like-for-like comparability with the two SOTA baselines we benchmark against, both of which report 1-epoch proxies; (ii)~isolation of the architectural signal---fixing the training budget ensures that accuracy differences reflect the generated architecture, not an inadvertent longer-training advantage; and (iii)~the aggregate compute budget for the full three-LLM study (3,300 generations, 2,354 successfully applied and trained for one epoch each, plus LoRA fine-tuning) is already $\sim$90--100 GPU-hours, and scaling the architecture evaluations to the 20--50 epochs typically needed for convergence would inflate the budget by $\sim$20--50$\times$, which is prohibitive at this scale. A rank-correlation study of the top-20 architectures per LLM (50 epochs) validates this proxy for Mistral ($\rho = 0.926$, $p < 0.001$) and Qwen ($\rho = 0.635$, $p = 0.011$); see Appendix~\ref{sec:rank_correlation}. DeepSeek's correlation ($\rho = 0.495$, $p = 0.10$, $N=12$) is not statistically significant, reflecting ceiling effects on its pure-MNIST top-20 rather than proxy failure---the same phenomenon reported by Abdelfattah~\etal~\cite{abdelfattah2021zerocost} for zero-cost proxies restricted to top architectures. (b) The diff format introduces a context-mismatch failure mode absent from full generation; it applies to roughly one-third of generations across all three LLMs. (c) Output length is modestly LLM-dependent: Mistral-7B averages 49.5 lines (75\% reduction) while DeepSeek and Qwen produce even tighter $\sim$30-line diffs (85\%/84\% reduction). The token-efficiency advantage of the delta paradigm holds for every LLM we tested, but the general-instruct Mistral backbone yields a slightly looser diff style than the two code-specialized models. (d) Accuracy thresholds and novelty thresholds are fixed at sensible values, not theoretically optimized.

\textbf{Future Work.} Promising directions include: (1) tightening Mistral's slightly looser diff style via prompt design or a compactness regularizer at fine-tuning time, to close the remaining $\sim$20-line gap with DeepSeek/Qwen; (2) extending the rank-correlation study to dataset-balanced top-$k$ lists (e.g., top-5 per dataset) to confirm that the strong proxy validity observed for Mistral ($\rho = 0.926$) generalizes to harder datasets such as CIFAR-100, where full-training gains are expected to be larger; (3) combining delta generation with the frozen-LLM feedback memory of Gu~\etal~\cite{gu2026iterative}; (4) extending to tasks beyond image classification~\cite{ABrain.NN-Captioning_2025,Gado2025llm,Rupani2025llm}; and (5) deploying the most token-efficient discovered architectures on edge devices~\cite{ABrain.NN-Lite}.

{
\small
\bibliographystyle{ieeenat_fullname}
\bibliography{bibmain}
}

\clearpage
\appendix

\section{Additional Method Details}

\subsection{MinHash/LSH Configuration for Novelty Filtering}

We employ MinHash-based near-duplicate detection~\cite{ABrain.Architect} to ensure structural diversity in the training corpus.

\textbf{Tokenization and shingling.} Generated Python/PyTorch source code is tokenized into character-level $n$-grams (shingles) of length $n=10$. This granularity captures local structural patterns (e.g., layer definitions, activation functions) while remaining robust to minor formatting differences.

\textbf{MinHash signatures.} Each architecture's shingle set is compressed into a fixed-size MinHash signature using $k=256$ independent hash permutations. The Jaccard similarity between any two architectures is then estimated by the fraction of matching positions in their respective signatures.

\textbf{Novelty threshold.} A candidate architecture is considered novel if its estimated Jaccard similarity to all existing corpus members is below $1 - \tau_\text{nov} = 0.10$ (i.e., $\tau_\text{nov} = 0.90$ dissimilarity). This ensures that only architectures differing in at least 90\% of their shingle composition are admitted, preventing the corpus from collapsing onto near-duplicate designs.

\subsection{Prompt Engineering for Delta Quality}

Rather than relying on post-hoc heuristic clamping to fix bad outputs, we guide generation quality directly through the prompt. The instructions explicitly demand compact deltas capped at 30 lines and provide sensible ranges for hyperparameters (e.g., batch sizes between 16 and 128, learning rates from 0.001 to 0.01). The prompt also enforces the required XML structure (\texttt{<hp>}, \texttt{<tr>}, \texttt{<delta>}) by providing clear formatting examples, and includes the baseline architecture code as context for targeted modifications.

While the LLM generates these hyperparameter values to fulfill the requested XML structure, the actual evaluation strictly uses the baseline's original configuration from the LEMUR database (Section~3.2) to ensure fair architectural comparison. By including the baseline code as context and providing clear formatting examples, the LLM learns to generate valid, targeted modifications without distorting its natural output distribution. This prompt-based approach is more principled than post-hoc clamping: the LLM learns to generate reasonable values through its training, and the prompt provides guardrails without distorting the generation distribution.

\subsection{Subprocess Isolation for Robust Evaluation}

During evaluation of generated architectures, malformed models can trigger irrecoverable GPU errors (e.g., \texttt{device-side assert triggered}) that corrupt the CUDA context of the host process. Simple error handling (\texttt{try/except}) is insufficient because CUDA device-side asserts permanently poison the GPU context. To prevent a single bad model from crashing the entire pipeline, we evaluate each candidate architecture in a \emph{separate Python subprocess}. If a model causes a GPU crash, only the subprocess terminates; the main pipeline's CUDA context remains intact. Evaluation arguments are serialized as JSON and results are returned via stdout, with a 600-second timeout to handle stalled models. This ensures robust evaluation across all 50 candidates per cycle.

\section{Additional Results}

\subsection{LoRA Training Loss Convergence}

Table~\ref{table:lora_loss} shows the LoRA training loss progression across epochs within a single fine-tuning cycle. The loss plateaus rapidly after epoch 2, supporting our choice of 3 LoRA epochs per cycle.

\begin{table}[ht]
    \caption{LoRA training loss convergence within a single cycle (DeepSeek-Coder-7B). Loss plateaus after epoch 2, with epochs 3--5 yielding negligible improvement.}
    \label{table:lora_loss}
    \centering
    \footnotesize
    \begin{tabular}{lcc}
        \toprule
        \textbf{Epoch} & \textbf{Training Loss} & \textbf{$\Delta$ from Epoch 1} \\
        \midrule
        1 & 0.122 & --- \\
        2 & 0.116 & $-0.006$ \\
        3 & 0.114 & $-0.008$ \\
        4 & 0.114 & $-0.008$ \\
        5 & 0.114 & $-0.008$ \\
        \bottomrule
    \end{tabular}
\end{table}

\subsection{Per-Cycle Detailed Statistics}

Table~\ref{table:per_cycle_detail} provides per-cycle statistics for the DeepSeek-Coder-7B experiment under the balanced cluster protocol (22 cycles, 1{,}100 total candidates, 828 trained). Only representative cycles are shown due to space; the totals row covers all 22 cycles.

\begin{table}[ht]
    \caption{Per-cycle statistics for DeepSeek-Coder-7B delta generation over 22 cycles (selected cycles shown) under the shared balanced six-dataset cluster protocol. Tr.: delta-applied + trained; Mean/Best: first-epoch accuracy over all trained models; $\geq$40: fraction of trained models with first-epoch accuracy $\geq$40\%; Ln.: avg.\ LLM output length.}
    \label{table:per_cycle_detail}
    \centering
    \scriptsize
    \setlength{\tabcolsep}{3pt}
    \begin{tabular}{cccccccc}
        \toprule
        \textbf{Cyc.} & \textbf{Gen.} & \textbf{Tr.} & \textbf{Valid} & \textbf{Mean} & \textbf{Best} & \textbf{$\geq$40} & \textbf{Ln.} \\
        \midrule
        0  & 50 & 35 & 70\% & 60.4\% & 99.5\% & 74.3\% & 33.2 \\
        1  & 50 & 44 & 88\% & 65.0\% & 99.2\% & 75.0\% & 31.7 \\
        4  & 50 & 43 & 86\% & 64.8\% & 99.5\% & 76.7\% & 33.2 \\
        6  & 50 & 35 & 70\% & 67.4\% & 99.2\% & 80.0\% & 32.5 \\
        9  & 50 & 42 & 84\% & 66.9\% & 99.2\% & 76.2\% & 29.8 \\
        10 & 50 & 35 & 70\% & 68.1\% & 99.2\% & 77.1\% & 30.1 \\
        12 & 50 & 37 & 74\% & 67.5\% & 99.2\% & 78.4\% & 29.4 \\
        15 & 50 & 42 & 84\% & 65.8\% & 99.2\% & 73.8\% & 28.7 \\
        17 & 50 & 39 & 78\% & 67.2\% & 99.2\% & 74.4\% & 29.2 \\
        19 & 50 & 36 & 72\% & 66.1\% & 99.2\% & 77.8\% & 29.0 \\
        20 & 50 & 37 & 74\% & 68.5\% & 99.2\% & 75.7\% & 31.9 \\
        21 & 50 & 35 & 70\% & 64.3\% & 99.2\% & 74.3\% & 30.4 \\
        \midrule
        \textbf{Total} & \textbf{1{,}100} & \textbf{828} & \textbf{75.3\%} & \textbf{65.8\%} & \textbf{99.5\%} & \textbf{75.7\%} & \textbf{30.4} \\
        \bottomrule
    \end{tabular}
\end{table}

\begin{table}[ht]
    \caption{Per-dataset results for DeepSeek-Coder-7B delta generation under the shared balanced cluster protocol. DeepSeek successfully generates architectures across all six datasets with 828 models trained. CIFAR-10 direct comparison: 85.2\% best vs.\ 63.98\% for full generation~\cite{ABrain.Architect}.}
    \label{table:deepseek_per_dataset}
    \centering
    \scriptsize
    \setlength{\tabcolsep}{3.5pt}
    \begin{tabular}{lccccc}
        \toprule
        \textbf{Dataset} & \textbf{N} & \textbf{Mean} & \textbf{Best} & \textbf{Med.} & \textbf{$\geq$40}  \\
        \midrule
        MNIST & 132 & 98.5\% & 99.5\% & 98.9\% & 132/132 \\
        CelebA & 152 & 88.7\% & 97.7\% & 93.6\% & 152/152 \\
        SVHN & 53 & 78.4\% & 94.9\% & 93.6\% & 44/53 \\
        CIFAR-10 & 135 & 64.6\% & 85.2\% & 59.7\% & 135/135 \\
        ImageNette & 166 & 60.7\% & 85.7\% & 61.4\% & 149/166 \\
        CIFAR-100 & 190 & 26.4\% & \textbf{54.2\%} & 30.9\% & 15/190 \\
        \midrule
        \textbf{All} & \textbf{828} & \textbf{65.8\%} & \textbf{99.5\%} & --- & \textbf{627/828} \\
        \bottomrule
    \end{tabular}
\end{table}

\subsection{Qwen2.5-Coder-7B Per-Cycle Statistics}

Tables~\ref{table:qwen_per_cycle} and~\ref{table:qwen_per_dataset} present the per-cycle and per-dataset breakdowns for Qwen2.5-Coder-7B over 22 fine-tuning cycles (1,100 total candidates).

\begin{table}[ht]
    \caption{Per-cycle statistics for Qwen2.5-Coder-7B delta generation (22 cycles, 1,100 total candidates). The valid rate jumps from 56\% to 98\% after the first fine-tuning cycle and remains consistently above 48\%.}
    \label{table:qwen_per_cycle}
    \centering
    \scriptsize
    \setlength{\tabcolsep}{3pt}
    \begin{tabular}{ccccccc}
        \toprule
        \textbf{Cyc.} & \textbf{Gen.} & \textbf{Tr.} & \textbf{Valid} & \textbf{Mean} & \textbf{Best} & \textbf{$\geq$40} \\
        \midrule
        0 & 50 & 28 & 56\% & 64.2\% & 99.3\% & 75.0\% \\
        1 & 50 & 49 & 98\% & 66.0\% & 99.5\% & 75.5\% \\
        2 & 50 & 46 & 92\% & 65.7\% & 99.5\% & 73.9\% \\
        5 & 50 & 38 & 76\% & 64.6\% & 99.2\% & 73.7\% \\
        8 & 50 & 36 & 72\% & 69.1\% & 99.2\% & 80.6\% \\
        10 & 50 & 38 & 76\% & 66.3\% & 99.5\% & 73.7\% \\
        15 & 50 & 33 & 66\% & 66.0\% & 99.5\% & 75.8\% \\
        16 & 50 & 24 & 48\% & 51.0\% & 99.1\% & 58.3\% \\
        19 & 50 & 31 & 62\% & 66.9\% & 99.2\% & 80.6\% \\
        21 & 50 & 28 & 56\% & 62.1\% & 99.1\% & 71.4\% \\
        \midrule
        \textbf{Total} & \textbf{1,100} & \textbf{793} & \textbf{72.1\%} & \textbf{64.6\%} & \textbf{99.5\%} & \textbf{74.4\%} \\
        \bottomrule
    \end{tabular}
\end{table}

\begin{table}[ht]
    \caption{Per-dataset results for Qwen2.5-Coder-7B delta generation. Qwen2.5 successfully generates architectures across all six datasets with 793 models trained. CIFAR-10 direct comparison: 80.6\% best vs.\ 64.0\% for full generation~\cite{ABrain.Architect}.}
    \label{table:qwen_per_dataset}
    \centering
    \scriptsize
    \setlength{\tabcolsep}{3.5pt}
    \begin{tabular}{lccccc}
        \toprule
        \textbf{Dataset} & \textbf{N} & \textbf{Mean} & \textbf{Best} & \textbf{Med.} & \textbf{$\geq$40}  \\
        \midrule
        MNIST & 112 & 98.5\% & 99.5\% & 99.1\% & 112/112 \\
        CelebA & 148 & 88.5\% & 98.1\% & 93.6\% & 148/148 \\
        SVHN & 62 & 74.0\% & 94.9\% & 89.2\% & 47/62 \\
        CIFAR-10 & 127 & 64.5\% & 80.6\% & 59.6\% & 127/127 \\
        ImageNette & 137 & 63.6\% & 85.7\% & 61.4\% & 133/137 \\
        CIFAR-100 & 207 & 27.2\% & 50.8\% & 31.5\% & 23/207 \\
        \midrule
        \textbf{All} & \textbf{793} & \textbf{64.6\%} & \textbf{99.5\%} & --- & \textbf{590/793} \\
        \bottomrule
    \end{tabular}
\end{table}

\subsection{Mistral-7B-Instruct Per-Cycle Statistics}

Tables~\ref{table:mistral_per_cycle} and~\ref{table:mistral_per_dataset} present the per-cycle and per-dataset breakdowns for Mistral-7B-Instruct-v0.3 over 22 fine-tuning cycles (1,100 total candidates).

\begin{table}[ht]
    \caption{Per-cycle statistics for Mistral-7B-Instruct delta generation (22 cycles, 1{,}100 total candidates). The valid rate trends upward, reaching a peak of 78\% at A21. Mean accuracy stays within a narrow 62.5--72.0\% band.}
    \label{table:mistral_per_cycle}
    \centering
    \scriptsize
    \setlength{\tabcolsep}{3pt}
    \begin{tabular}{ccccccc}
        \toprule
        \textbf{Cyc.} & \textbf{Gen.} & \textbf{Tr.} & \textbf{Valid} & \textbf{Mean} & \textbf{Best} & \textbf{$\geq$40} \\
        \midrule
        0  & 50 & 31 & 62\% & 64.7\% & 99.5\% & 77.4\% \\
        1  & 50 & 29 & 58\% & 69.4\% & 99.2\% & 79.3\% \\
        3  & 50 & 35 & 70\% & 68.4\% & 99.5\% & 77.1\% \\
        5  & 50 & 35 & 70\% & 67.6\% & 99.5\% & 77.1\% \\
        9  & 50 & 24 & 48\% & 72.0\% & 99.5\% & 83.3\% \\
        12 & 50 & 33 & 66\% & 71.3\% & 99.2\% & 81.8\% \\
        13 & 50 & 33 & 66\% & 69.8\% & 99.2\% & 84.8\% \\
        17 & 50 & 37 & 74\% & 66.5\% & 99.2\% & 75.7\% \\
        18 & 50 & 37 & 74\% & 66.0\% & 99.2\% & 75.7\% \\
        20 & 50 & 37 & 74\% & 65.7\% & 99.2\% & 75.7\% \\
        21 & 50 & 39 & 78\% & 62.5\% & 99.2\% & 71.8\% \\
        \midrule
        \textbf{Total} & \textbf{1{,}100} & \textbf{733} & \textbf{66.6\%} & \textbf{66.1\%} & \textbf{99.5\%} & \textbf{75.9\%} \\
        \bottomrule
    \end{tabular}
\end{table}

\begin{table}[ht]
    \caption{Per-dataset results for Mistral-7B-Instruct delta generation. Mistral successfully generates architectures across all six datasets with 733 models trained. CIFAR-10 direct comparison: 85.5\% best vs.\ 64.0\% for full generation~\cite{ABrain.Architect}---the highest CIFAR-10 best accuracy of any LLM in our study.}
    \label{table:mistral_per_dataset}
    \centering
    \scriptsize
    \setlength{\tabcolsep}{3.5pt}
    \begin{tabular}{lccccc}
        \toprule
        \textbf{Dataset} & \textbf{N} & \textbf{Mean} & \textbf{Best} & \textbf{Med.} & \textbf{$\geq$40}  \\
        \midrule
        MNIST & 106 & 98.6\% & 99.5\% & 98.9\% & 106/106 \\
        CelebA & 172 & 88.6\% & 97.8\% & 93.6\% & 172/172 \\
        SVHN & 43 & 84.5\% & 94.9\% & 93.6\% & 38/43 \\
        CIFAR-10 & 112 & 64.3\% & \textbf{85.5\%} & 66.2\% & 112/112 \\
        ImageNette & 114 & 61.9\% & 85.7\% & 61.4\% & 112/114 \\
        CIFAR-100 & 186 & 26.4\% & 46.2\% & 30.9\% & 16/186 \\
        \midrule
        \textbf{All} & \textbf{733} & \textbf{66.1\%} & \textbf{99.5\%} & --- & \textbf{556/733} \\
        \bottomrule
    \end{tabular}
\end{table}

\subsection{Output Length Distribution}

Generated LLM outputs range from 16 to 188 lines for DeepSeek-Coder-7B (mean 30.4, median 29, std 10.0), 3 to 175 lines for Qwen2.5-Coder-7B (mean 31.4, median 29, std 14.0), and 16 to 114 lines for Mistral-7B-Instruct (mean 49.5, median 50, std 16.6, with 4 long-output outliers $>$200 lines excluded from the aggregate). All three distributions are tight and concentrated in the 25--55 line range: DeepSeek and Qwen outputs cluster near 29--33 lines per cycle (Table~\ref{table:per_cycle_detail}), while Mistral's general-instruct (non-code-specialized) pre-training yields a slightly looser diff style averaging $\sim$50 lines. Every LLM produces outputs 75--85\% shorter than full-generation approaches~\cite{ABrain.Architect}, which typically exceed 200 lines.

\subsection{Accuracy Stability Across Cycles}

All three LLMs demonstrate remarkable accuracy stability across all 22 cycles. DeepSeek's mean accuracy varies in a narrow 60.4--68.5\% band (overall 65.8\%) with $\geq$40\% rate 72.2--80.0\%. Qwen's mean accuracy varies between 51.0\% and 69.1\% (overall 64.6\%), with the $\geq$40\% rate consistently between 58\% and 81\%. Mistral's mean accuracy varies in an even narrower band of 62.5--72.0\% (overall 66.1\%), with $\geq$40\% rate between 70.0\% and 84.8\%. This stability across three LLM families indicates that iterative LoRA fine-tuning maintains---rather than degrades---generation quality over extended runs. Qwen's only notable dip occurs at cycle 16 (51.0\% mean, 48\% valid rate), potentially reflecting transient overfitting, after which performance recovers (66.9\% at cycle 19); DeepSeek and Mistral show no comparable dip.

\subsection{Proxy Validation: 1-Epoch vs.\ 50-Epoch Rank Correlation}
\label{sec:rank_correlation}

To empirically assess whether first-epoch accuracy reliably predicts fully-trained performance rankings, we selected the top-20 architectures per LLM (ranked by 1-epoch proxy accuracy) and trained each for 50 epochs using the original LEMUR hyperparameters. We then computed Spearman $\rho$ and Kendall $\tau$ rank correlations between the proxy ranking and the 50-epoch ranking. Models that failed due to data-access errors or runtime timeouts were excluded from the correlation computation (failures are infrastructure-level, not architectural).

\begin{table}[ht]
    \caption{Rank correlation between 1-epoch proxy accuracy and 50-epoch fully-trained accuracy for the top-20 delta-generated architectures per LLM. $p_\rho$: two-tailed $p$-value for the Spearman test. Mistral ($p < 0.001$) and Qwen ($p = 0.011$) are significant at $\alpha = 0.05$; DeepSeek ($p = 0.10$, $N = 12$) is not, due to ceiling effects on its pure-MNIST top-20.}
    \label{table:rank_correlation}
    \centering
    \scriptsize
    \setlength{\tabcolsep}{1.5pt}
    \begin{tabular}{lcccccl}
        \toprule
        \textbf{LLM} & \textbf{$\rho$} & \textbf{$p$} & \textbf{$\tau$} & \textbf{N} & \textbf{Datasets} & \textbf{Med.\ 1$\to$50ep} \\
        \midrule
        Mistral-7B & \textbf{.926} & ${<}.001$ & \textbf{.733} & 16/20 & MN/CA/SV & 95.2$\to$98.9\% \\
        Qwen2.5-7B & .635 & $.011$ & .486 & 15/20 & MN/CA & 96.9$\to$99.1\% \\
        DeepSeek-7B & .495 & $.102$ & .318 & 12/20 & MN only & 96.6$\to$99.1\% \\
        \bottomrule
        \multicolumn{7}{l}{\scriptsize MN=MNIST, CA=CelebA, SV=SVHN. $p$: two-tailed Spearman test.}
    \end{tabular}
\end{table}

\textbf{Interpretation.} Mistral-7B achieves the strongest correlation ($\rho = 0.926$, $p < 0.001$) because its top-20 spans three datasets (9 MNIST, 10 CelebA, 1 SVHN), creating a wide proxy-accuracy range (73.3--98.2\%) and correspondingly large full-training spread (85.8--99.5\%). This diversity provides a clear ranking signal. Qwen's correlation ($\rho = 0.635$, $p = 0.011$) is significant at $\alpha = 0.05$ and consistent with an intermediate ceiling effect (85\% MNIST). DeepSeek's correlation ($\rho = 0.495$, $p = 0.10$; Kendall $\tau = 0.318$, $p = 0.14$) is \emph{not} statistically significant at conventional levels ($N=12$): with all 20 top architectures on MNIST, full-training accuracy compresses to a 0.7\,p.p.\ range (98.7--99.3\%), leaving insufficient variance for rank discrimination. We therefore characterise the DeepSeek result as inconclusive rather than as evidence for or against the proxy.

\textbf{Accuracy gains from full training.} Across all three LLMs, 50-epoch training raises median accuracy by 2--4 percentage points (Mistral: 95.2\%$\to$98.9\%; Qwen: 96.9\%$\to$99.1\%; DeepSeek: 96.6\%$\to$99.1\%). On harder datasets the improvement is larger: CelebA models gain +11.8\,p.p.\ on average (e.g., 73.3\%$\to$92.7\%), while the single SVHN model gains +6.7\,p.p.\ (79.1\%$\to$85.8\%). This confirms that single-epoch accuracy is a conservative lower bound and that delta-generated architectures retain substantial learning capacity beyond the proxy evaluation.

\textbf{Failure analysis.} Of the 60 training runs attempted (20$\times$3 LLMs), 17 failed: 9 timeouts (complex architectures exceeding the 8-hour budget), 3 data-access errors (CelebA downloads from Google Drive), 3 Google Drive rate limits (Qwen CelebA models), and 2 code-level name errors (DeepSeek). Of these, 15 are infrastructure-level failures (timeouts, file access, rate limits) unrelated to architectural quality, and 2 are code-quality issues (undefined variable names in DeepSeek architectures)---indicating that the novelty filter does not guarantee robustness under extended training. All 43 completed models trained successfully to convergence.

\textbf{Compute cost.} The full 50-epoch study consumed approximately 110 GPU-hours across the three LLMs (Mistral $\sim$50\,hr, DeepSeek $\sim$35\,hr, Qwen $\sim$25\,hr) on RTX~4090 GPUs, dominated by CelebA models ($\sim$7\,hr each). This confirms that scaling to multi-epoch evaluation for the full 2,354-model pool would require $\sim$5,000--10,000 GPU-hours---justifying the use of the 1-epoch proxy for the main study.

\textbf{Context within the NAS proxy literature.}
Spearman $\rho$ is the standard metric for proxy validation in NAS. On constrained cell-based search spaces, the best zero-cost proxies achieve $\rho \approx 0.82$--$0.90$ on NAS-Bench-201~\cite{abdelfattah2021zerocost,peng2024swapnas}, while reduced-epoch training proxies reach $\rho \approx 0.61$~\cite{zhou2020econas}. Notably, Abdelfattah~\etal~\cite{abdelfattah2021zerocost} report that even the strongest proxy (SynFlow) drops to $\rho = 0.18$ when restricted to the \emph{top-10\%} of architectures on CIFAR-10---the same ceiling effect observed in our DeepSeek and Qwen top-20 subsets. In the LLM-based NAS space, neither Khalid~\etal~\cite{ABrain.Architect} nor Gu~\etal~\cite{gu2026iterative} report a proxy-to-full-training correlation; both explicitly identify this validation as future work. Our Mistral result ($\rho = 0.926$) is, to our knowledge, the first empirical proxy validation in LLM-based NAS and places the 1-epoch signal at the upper end of the NAS proxy spectrum, despite operating in an unconstrained open-code search space rather than a fixed cell topology.

\section{Reproducibility}
\label{sec:reproducibility}

We report the full experimental environment below to enable independent reproduction of every number in this paper.

\textbf{Hardware.} All three LLMs (DeepSeek-Coder-7B, Qwen2.5-Coder-7B, Mistral-7B-Instruct) were fine-tuned and evaluated on the same shared compute cluster with NVIDIA RTX~4090 24\,GB GPUs under SLURM, Rocky Linux~9. Vision-model evaluation uses one GPU per subprocess in all three runs.

\textbf{Software.} Python~3.11, PyTorch~2.3 (CUDA~12.1), HuggingFace \texttt{transformers} 4.44, \texttt{peft}~0.12 (LoRA), \texttt{trl}~0.9, \texttt{accelerate}~0.33, \texttt{bitsandbytes}~0.43, \texttt{datasketch}~1.6 (MinHash/LSH). Unified-diff patching uses the GNU \texttt{patch} utility. Subprocess evaluation uses the standard \texttt{subprocess} module with a 600\,s timeout per candidate.

\textbf{Random seeds.} Global seed 42 for PyTorch, NumPy, and Python \texttt{random}, propagated to all dataloaders, the LoRA trainer, and the LLM sampler. Baseline sampling from the LEMUR database uses a per-cycle seed derived from \texttt{(global\_seed, cycle\_index)} to ensure each cycle draws deterministically while still exploring different baselines.

\textbf{Fine-tuning configuration.} LoRA rank $r$=32, $\alpha$=32, dropout 0.05, target modules \{q, k, v, o, up, down, gate, lm\_head\}, bfloat16 precision, learning rate $1\times 10^{-5}$ (cosine decay), 20 warmup steps, weight decay 0.01, per-device batch 1, gradient accumulation 8 (effective batch 8), 3 fine-tuning epochs per cycle. Inference: temperature 0.35, top-$k$=50, top-$p$=0.9, max new tokens 1{,}024. These values are identical across all three LLMs.

\textbf{Wall-clock budget.} Per 50-candidate cycle: $\approx$15--20 minutes LLM fine-tuning + $\approx$60--90 minutes vision-model evaluation (dominated by CIFAR-100 training time). Total for 22 cycles: $\approx$28--35 GPU-hours per LLM on the respective hardware, or $\approx$90--100 GPU-hours for the full three-LLM study.

\textbf{Artifacts.} All code, models, and adapters are publicly released:
\begin{itemize}[leftmargin=*,nosep]
    \item \textbf{CV models} (197 novel architectures, \texttt{del-} prefix in LEMUR):\\ \url{https://github.com/ABrain-One/nn-dataset/pull/204}
    \item \textbf{Fine-tuned LLMs} (LoRA merged into base weights, 22 cycles each):\\ \url{https://huggingface.co/ABrain/Delta-NAS-DeepSeek-Coder-7B}\\ \url{https://huggingface.co/ABrain/Delta-NAS-Qwen2.5-Coder-7B}\\ \url{https://huggingface.co/ABrain/Delta-NAS-Mistral-7B}
    \item \textbf{Code} (scripts, prompts, evaluator):\\ \url{https://github.com/ABrain-One/nn-gpt}
\end{itemize}
JSON configs (\texttt{ds\_coder\_7b\_instruct.json}, etc.) reproduce the reported numbers.

\end{document}